\documentclass[11pt,letterpaper]{article}

\usepackage[letterpaper,left=1.05in,right=1.05in,top=1in,bottom=1in]{geometry}
\usepackage[T1]{fontenc}
\usepackage[utf8]{inputenc}
\usepackage[english]{babel}

\usepackage{mathpazo}
\usepackage{tgpagella}
\usepackage{inconsolata}
\usepackage{microtype}
\usepackage{amsmath,amsfonts,amssymb,mathtools}
\usepackage{graphicx}
\usepackage{booktabs}
\usepackage{multirow}
\usepackage{tabularx}
\usepackage{array}
\usepackage{wrapfig}
\usepackage{nicefrac}
\usepackage[table,dvipsnames]{xcolor}
\usepackage{pifont}
\usepackage{fontawesome5}
\usepackage{tikz}
\usetikzlibrary{positioning,arrows.meta,fit,backgrounds,decorations.pathmorphing,calc}
\usepackage{tcolorbox}
\tcbuselibrary{skins,breakable}
\usepackage{enumitem}
\usepackage{placeins}
\usepackage{caption}
\usepackage{subcaption}
\usepackage{authblk}
\usepackage[explicit]{titlesec}
\usepackage{fancyhdr}
\usepackage{lastpage}

\PassOptionsToPackage{authoryear,round}{natbib}
\usepackage{natbib}

\usepackage[bookmarks=true,bookmarksopen=true]{hyperref}

\definecolor{inkblack}  {HTML}{0A0A0A}   
\definecolor{inkdark}   {HTML}{1A1A1A}   
\definecolor{inkmid}    {HTML}{2B2B2B}   
\definecolor{inkgrey}   {HTML}{4A4A4A}   
\definecolor{inkmuted}  {HTML}{6E6E6E}   
\definecolor{inkfaint}  {HTML}{8C8C8C}   
\definecolor{inkrule}   {HTML}{BFBFBF}   
\definecolor{inkbg}     {HTML}{F7F7F7}   
\definecolor{inkbg2}    {HTML}{F0F0F0}   
\definecolor{inkframe}  {HTML}{D9D9D9}   

\colorlet{narrared}{inkmid}
\colorlet{narrareddeep}{inkblack}
\colorlet{cforge}{inkdark}
\colorlet{cmemory}{inkgrey}
\colorlet{cpace}{inkmid}
\colorlet{cmeta}{inkmuted}
\colorlet{cnovelty}{inkdark}
\definecolor{ctablehead}{HTML}{D6D6D6}
\definecolor{ctablerule}{HTML}{BFBFBF}
\colorlet{ctabletext}{inkdark}
\colorlet{cstoryq}{inkblack}
\colorlet{cux}{inkgrey}
\colorlet{openyes}{inkblack}
\colorlet{closedno}{inkfaint}

\hypersetup{
    colorlinks  = true,
    linkcolor   = inkblack,
    citecolor   = inkmid,
    urlcolor    = inkmid,
    filecolor   = inkblack,
    pdfborder   = {0 0 0}
}

\titleformat{\section}
  {\Large\bfseries\color{inkblack}}{\thesection}{0.6em}{#1}[]
\titleformat{\subsection}
  {\large\bfseries\color{inkmid}}{\thesubsection}{0.55em}{#1}[]
\titleformat{\subsubsection}
  {\normalsize\bfseries\itshape\color{inkmid}}{\thesubsubsection}{0.5em}{#1}[]
\titlespacing*{\section}      {0pt}{2.6ex plus 0.8ex minus 0.4ex}{1.0ex}
\titlespacing*{\subsection}   {0pt}{2.0ex plus 0.7ex minus 0.3ex}{0.8ex}
\titlespacing*{\subsubsection}{0pt}{1.5ex plus 0.5ex minus 0.2ex}{0.6ex}

\setlist[itemize,1]{label=\textbullet,leftmargin=1.4em,topsep=2pt,itemsep=2pt,parsep=0pt}
\setlist[itemize,2]{label=$\circ$,leftmargin=1.2em}
\setlist[enumerate,1]{leftmargin=1.6em,topsep=2pt,itemsep=2pt,parsep=0pt}

\renewcommand\Affilfont{\small\color{inkmuted}}
\makeatletter
\renewcommand\AB@affilsepx{,\enspace\protect\Affilfont}
\makeatother

\fancypagestyle{firststyle}{%
  \fancyhf{}%
  \fancyfoot[C]{\footnotesize\color{inkmuted}\thepage\ /\ \pageref{LastPage}}%
}

\newcommand{\narraemph}[1]{\textcolor{narrareddeep}{\textbf{#1}}}
\newcommand{\cmark}[2]{\textcolor{#1}{\textbf{#2}}}
\newcommand{\narraforge}{\cmark{cforge}{Narrative Architect}}
\newcommand{\memorylens}{\cmark{cmemory}{Memory Agent}}
\newcommand{\pacekeeper}{\cmark{cpace}{Pacing Agent}}
\newcommand{\metaplanner}{\cmark{cmeta}{Planning Agent}}
\newcommand{\noveltygate}{\cmark{cnovelty}{Artifact Agent}}

\newcommand{\phead}[1]{\noindent\textcolor{narrareddeep}{\textbf{#1}}\ }

\newenvironment{compactlist}{%
  \begin{list}{\textcolor{narrared}{\textbullet}}{%
    \setlength{\leftmargin}{0pt}%
    \setlength{\itemindent}{1.1em}%
    \setlength{\itemsep}{2pt}%
    \setlength{\parsep}{0pt}%
    \setlength{\topsep}{4pt}%
  }%
}{\end{list}}
\newenvironment{numcompactlist}{%
  \begin{list}{}{%
    \setlength{\leftmargin}{0pt}%
    \setlength{\itemindent}{1.7em}%
    \setlength{\labelwidth}{1.2em}%
    \setlength{\labelsep}{0.45em}%
    \setlength{\itemsep}{2pt}%
    \setlength{\parsep}{0pt}%
    \setlength{\topsep}{4pt}%
  }%
}{\end{list}}

\newcommand{\motivbox}[1]{%
  \par\medskip
  \noindent
  \begin{tikzpicture}
    \node[inner xsep=10pt, inner ysep=8pt, text width=0.93\linewidth,
          fill=inkbg!90, rounded corners=2pt, anchor=west] (box)
      {\small\color{inkdark}\itshape #1};
    \draw[inkblack!85, line width=1.3pt] (box.south west) -- (box.north west);
  \end{tikzpicture}
  \par\medskip
}

\newcommand{\cnum}[1]{\ding{\the\numexpr181+#1\relax}}

\title{\textsc{NARRA-Gym} for Evaluating Interactive Narrative Agents}

\author[1]{Yue Huang\textsuperscript{\textdagger}}
\author[2,11]{Yuchen Ma\textsuperscript{\textdagger,\textdaggerdbl}}
\author[3]{Jiayi Ye\textsuperscript{\textdagger}}
\author[1]{Wenjie Wang}
\author[4]{Zipeng Ling}
\author[5]{Xingjian Hu}
\author[6]{Yuexing Hao}
\author[7,8,9]{Zichen Chen}
\author[10]{Zhangchen Xu}
\author[1]{Yunhong He}
\author[1]{Zhengqing Yuan}
\author[1]{Yujun Zhou}
\author[1]{Kehan Guo}
\author[1]{Chaoran Chen}
\author[1]{Toby Jia-Jun Li}
\author[2,11]{Stefan Feuerriegel}
\author[1]{Xiangliang Zhang\textsuperscript{*}}

\affil[1]{University of Notre Dame}
\affil[2]{LMU Munich}
\affil[3]{Independent Researcher}
\affil[4]{University of Pennsylvania}
\affil[5]{Lehigh University}
\affil[6]{Massachusetts Institute of Technology}
\affil[7]{Bake AI}
\affil[8]{UC Santa Barbara}
\affil[9]{Stanford University}
\affil[10]{University of Washington}
\affil[11]{Munich Center for Machine Learning}

\newcommand{\equalcontributionnote}{%
  \textsuperscript{\textdagger}\,These authors contributed equally to this work.\quad
  \textsuperscript{*}\,Corresponding author: \texttt{xzhang33@nd.edu}.\\[2pt]
  \textsuperscript{\textdaggerdbl}\,Yuchen Ma is supported by the DAAD program ``Konrad Zuse Schools of Excellence in Artificial Intelligence,'' sponsored by the Federal Ministry of Education and Research.}

\newcommand{\paperstatus}{Preprint. Under review.}
\newcommand{\paperurl}{https://github.com/narra-gym/NARRA-Gym}
\newcommand{\projecturl}{https://narra-gym.github.io/}

\makeatletter
\newcommand{\makecoverpage}{%
  \thispagestyle{firststyle}%
  \vspace*{-1.0em}%
  \noindent
  {\sffamily\bfseries\color{inkblack}\large NARRA-Gym}\hspace{0.6em}%
  {\sffamily\color{inkmuted}\small Technical Report}\par
  \vspace{4pt}
  {\color{inkblack}\rule{\linewidth}{1.2pt}}\par
  \vspace{1.6em}

  {\raggedright\bfseries\color{inkblack}\fontsize{20pt}{25pt}\selectfont
   \@title\par}
  \vspace{1.0em}

  {\raggedright\@author\par}
  \vspace{0.35em}
  {\itshape\color{inkmuted}\paperstatus\par}
  \vspace{0.25em}
  {\color{inkmid}%
   \faGithub\ \href{\paperurl}{\paperurl}\hspace{1.4em}%
   \faGlobe\ \href{\projecturl}{\projecturl}\par}
  \ifx\equalcontributionnote\@empty\else
    \vspace{0.15em}%
    {\footnotesize\itshape\color{inkmuted}\equalcontributionnote\par}%
  \fi
  \vspace{1.2em}

  \begin{tcolorbox}[
      enhanced, sharp corners, width=\textwidth,
      colback=inkbg, colframe=inkbg, boxrule=0pt,
      left=16pt,right=16pt,top=12pt,bottom=12pt
  ]
    {\bfseries\color{inkblack}\large Abstract}\par
    \vspace{0.45em}
    {\small\color{inkdark}\theabstract\par}
  \end{tcolorbox}
  \vspace{0.8em}
}
\makeatother

\newcommand{\theabstract}{%
Interactive narrative tasks require LLMs to sustain a coherent, evolving story while adapting to a user over multiple turns. However, suitable benchmarks for this setting are limited: existing evaluations often focus on static prompts, isolated story generations, or post-hoc ratings, and therefore miss whether models can jointly manage story generation, long-context state and pacing, character simulation, empathic personalization, and story-grounded artifacts. We introduce \textsc{NARRA-Gym}, an executable evaluation environment that turns a sparse emotional seed into a complete interactive story episode and logs the full model-in-the-loop trajectory, including story construction, memory updates, planning, pacing interventions, and optional artifact synthesis. We evaluate nine frontier LLMs using a controlled LLM-as-judge sweep over eight benchmark personas and a human evaluation in which participants rate customized model outputs. Our results show substantial variation across models, personas, and evaluation dimensions: models that produce fluent stories can still fail on robustness, user experience, or resistance-sensitive personalization. These findings suggest that interactive narrative offers a useful benchmark for evaluating long-horizon, user-adaptive LLM behavior beyond isolated story quality.%
}

\begin{document}

\makecoverpage

\section{Introduction}
\label{sec:introduction}

Interactive narrative refers to settings in which an LLM must sustain a coherent, evolving story world while interacting with a user over multiple turns by adapting both the narrative and the behavior in response to user input \citep{urbanek2019light,akoury-etal-2020-storium,du-chilton-2023-storywars,park2023generativeagents}. Such capabilities are increasingly used in creative domains, including collaborative storytelling, games, and interactive media, where models act as live narrative agents.\footnote{For example, game companies are already experimenting with generative models for live characters and narrative production. Examples are Ubisoft's NEO non-player character (NPC) prototype, Xbox--Inworld narrative tools, and NVIDIA ACE for natural-language NPC interaction \citep{ubisoft2024neonpc,xbox2023inworld,nvidia2023ace}.} At the same time, interactive narrative provides a challenging testbed for language models more broadly, because it requires a combination of multiple capabilities (i.e., generation, memory, planning, and personalization) under continuous, multi-turn interaction.

The above task is challenging because it extends far beyond conventional story writing. Unlike static text generation \citep{fan2018hierarchical,yao2019planwrite,guan-etal-2022-lot}, this requires the LLM to manage story progression, character consistency, a long-context state, and user alignment over multiple turns; simulate consistent characters; and adapt to the user's emotional trajectory. For example, the LLM must introduce new events while preserving prior context, keep characters psychologically coherent, respond appropriately to shifting user signals, and, in some cases, externalize the story through interactive artifacts such as letters, maps, or small interfaces. Together, this makes interactive narrative a difficult benchmarking setting, where failures in memory, planning, or alignment can break the interaction even when individual responses appear fluent.

Existing benchmarks for evaluating interactive narrative have key shortcomings. Standard LLM benchmarks emphasize static, \textit{single-turn} tasks such as question answering and closed-form reasoning \citep{hendrycks2021mmlu,bigbench2022}. Story-centric resources and surveys have broadened the evaluation to include narrative generation and narrative understanding \citep{guan-etal-2022-lot,yang2024storyevalsurvey,wang2023openworldsurvey,zhu2023narrativeunderstanding}, but many protocols still evaluate isolated generations, offline corpora, or post-hoc ratings. Hence, the benchmarks typically capture whether an LLM writes a plausible passage, but the benchmark misses whether the LLM can manage a long interactive narrative by keeping the story, cast, user history, and emotional contract coherent across turns.

Interactive narrative therefore presents a challenging dynamic testbed for LLMs; it requires an interplay of capabilities that are often evaluated separately to operate together: planning must survive improvisation, memory must support character consistency, empathy must shape story direction, and generated artifacts must remain grounded in the evolving fiction. Hence, a failure in any one of these abilities can break the entire interaction, \textit{even} if each individual response sounds fluent. Here, we introduce \textsc{NARRA-Gym}, an executable evaluation environment for benchmarking interactive narrative agents over multi-turn interaction. 

\textsc{NARRA-Gym} is motivated by \textbf{five coupled capabilities} that only fully surface under sustained interaction:
\begin{numcompactlist}
\item[\textcolor{narrareddeep}{\cnum{1}}] \textbf{Creative story generation.}
The model must construct a complete narrative arc from a sparse emotional seed, requiring both compelling prose and high-level story steering \citep{fan2018hierarchical,yao2019planwrite,wang2024dome,bae2024collective,gomez-rodriguez-williams-2023-confederacy}.

\item[\textcolor{narrareddeep}{\cnum{2}}] \textbf{Long-context state and pacing management.}
The model must keep dialogue history, unresolved tensions, revealed clues, user decisions, and the current narrative tempo available as actionable context without contradiction, drift, or stagnation \citep{liu2024lostmiddle,bai2024longbench,lyu2024facttrack,wu2025longeval}.

\item[\textcolor{narrareddeep}{\cnum{3}}] \textbf{Character simulation.}
Characters must remain distinguishable in voice and motivation while evolving with the plot \citep{park2023generativeagents,wang2024rolellm,han-etal-2024-ibsen,chen2024hollmwood,papoudakis-etal-2024-bookworm}.

\item[\textcolor{narrareddeep}{\cnum{4}}] \textbf{Empathic personalization.}
The story must align with the user's emotional needs without collapsing into generic therapeutic phrasing \citep{rashkin2019empathetic,harel-canada-etal-2024-measuring,yunusov2024mirrorstories}.

\item[\textcolor{narrareddeep}{\cnum{5}}] \textbf{Interactive artifact generation.}
The model must produce functional, story-grounded HTML, CSS, and JavaScript artifacts that remain novel and integrated with the evolving narrative \citep{urbanek2019light,akoury-etal-2020-storium,yang2024seedstory}.
\end{numcompactlist}


Following the environment-based evaluation framing of OpenAI Gym \citep{brockman2016openai}, \textsc{NARRA-Gym} places each tested model inside the same repeatable episode scaffold, where each generated response updates the next state. The scaffold is not the object being ranked; it is rather the controlled interaction setting that makes model differences observable by keeping the session runnable, logged, and comparable. An episode begins from a sparse \emph{emotional seed}, constructs a structured story world, and then runs a multi-turn interaction loop with logged memory updates, pacing checks, planning traces, and optional story-grounded artifacts. This design turns interactive storytelling from a loosely specified demo setting into a reproducible evaluation protocol.

Our \textbf{contributions} are:
\begin{numcompactlist}
\item[\textcolor{narrareddeep}{\cnum{1}}] \textbf{An executable benchmark environment.}
We define an interactive evaluation setting that jointly tests creative story generation, long-context state and pacing management, character simulation, empathic personalization, and story-grounded artifact generation inside a single interaction loop.

\item[\textcolor{narrareddeep}{\cnum{2}}] \textbf{A modular narrative-agent pipeline.}
We implement a staged story construction pipeline, including multi-resolution memory, reflection-guided planning, anti-stagnation control, novelty-constrained artifact synthesis, and fail-soft structured generation, where each component is logged for inspection.

\item[\textcolor{narrareddeep}{\cnum{3}}] \textbf{A comparative evaluation protocol.}
We provide a human rating protocol with within-group rank aggregation, together with LLM-judge protocols for comparing generator models across personas, rubric dimensions, and judge calibrations, exposing failures that are difficult to observe in static narrative datasets.
\end{numcompactlist}

\section{Evaluation Environment Construction}
\label{sec:environment}

\begin{figure}
    \centering
    \includegraphics[width=1\linewidth]{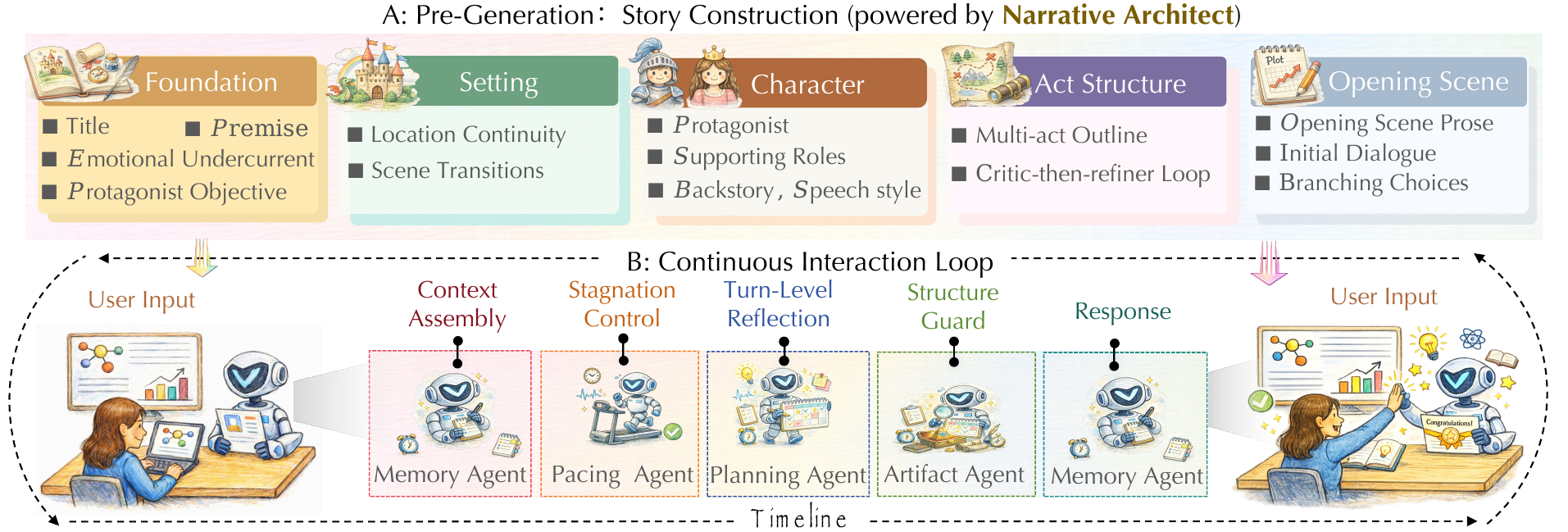}
    \vspace{-15pt}
    \caption{\textbf{Pipeline view of a \textsc{NARRA-Gym} episode.} \textbf{(A)} The \narraforge{} turns the user's emotional seed into the basic elements of a structured story world through five construction stages: story foundation, setting construction, character construction, act structure, and opening scene generation. \textbf{(B)} The initialized story then enters a continuous interaction loop coordinated by the remaining four agents: user choice or free-text input, context assembly by \memorylens{}, LLM story advancement, pacing and structure checks by \pacekeeper{}, optional planning by \metaplanner{}, optional artifact generation by \noveltygate{}, and state/log updates before the next turn.}
    \label{fig:architecture}
\vspace{-15pt}
\end{figure}

\textsc{NARRA-Gym} orchestrates the model through a complete episode pipeline, summarized in Figure~\ref{fig:architecture}. Here, an \emph{episode} means one full interactive story session from initial user input to the final logged conversation. The episode begins with the user's \emph{emotional seed}: a free-text description of their current situation or mood, entered through the start interface shown in Figure~\ref{fig:start} (Appendix~\ref{app:screenshots}). This seed can be enriched by \emph{profiling answers} (short questionnaire responses about preferences and comfort boundaries) and \emph{selected keywords} (user-chosen descriptors that should influence the story), as illustrated in Figure~\ref{fig:profiling}. The \narraforge{} then converts this sparse input into a runnable story world through five logged construction stages: (1)~story foundation, (2)~setting construction, (3)~character construction, (4)~act structure, and (5)~opening scene generation. The output of this initialization phase is not just prose, but a \emph{structured episode state}: machine-readable fields for the premise, setting, cast, act outline, opening dialogue, hidden elements, and initial choices. Figure~\ref{fig:story-cast} shows a representative generated synopsis and cast view from this construction phase.

After initialization, the episode enters a \emph{turn-level interaction loop}, meaning the repeated cycle that runs after each user action. The user can either select a displayed choice or type a \emph{free-form message}, an open text input that is not limited to the displayed choices. For each turn, a \memorylens{} assembles recent dialogue, profile information, \emph{story memory} (persistent fields such as current goal, clues, and tensions), and \emph{user-journey state} (the user's recorded decisions and emotional trajectory); the LLM generates the next story beat; a \pacekeeper{} and the structure guard check whether the plot actually advanced; a \metaplanner{} optionally produces planning guidance for subsequent turns; and an \noveltygate{} can generate story-grounded interactive artifacts when the narrative calls for a tangible prop. The response, choices, memory updates, pacing interventions, artifact metadata, and LLM traces are written back into the episode state before the next user action.

\subsection{From User Input to Story World}

\motivbox{Before any interaction begins, the system must turn sparse emotional input into a fully realized narrative world. The \narraforge{} decomposes this into five logged stages so that, for example, a model that produces good premises but flat characters can be distinguished from one that fails at act-level planning.}

\begin{figure}[t]
\centering
\includegraphics[width=\linewidth]{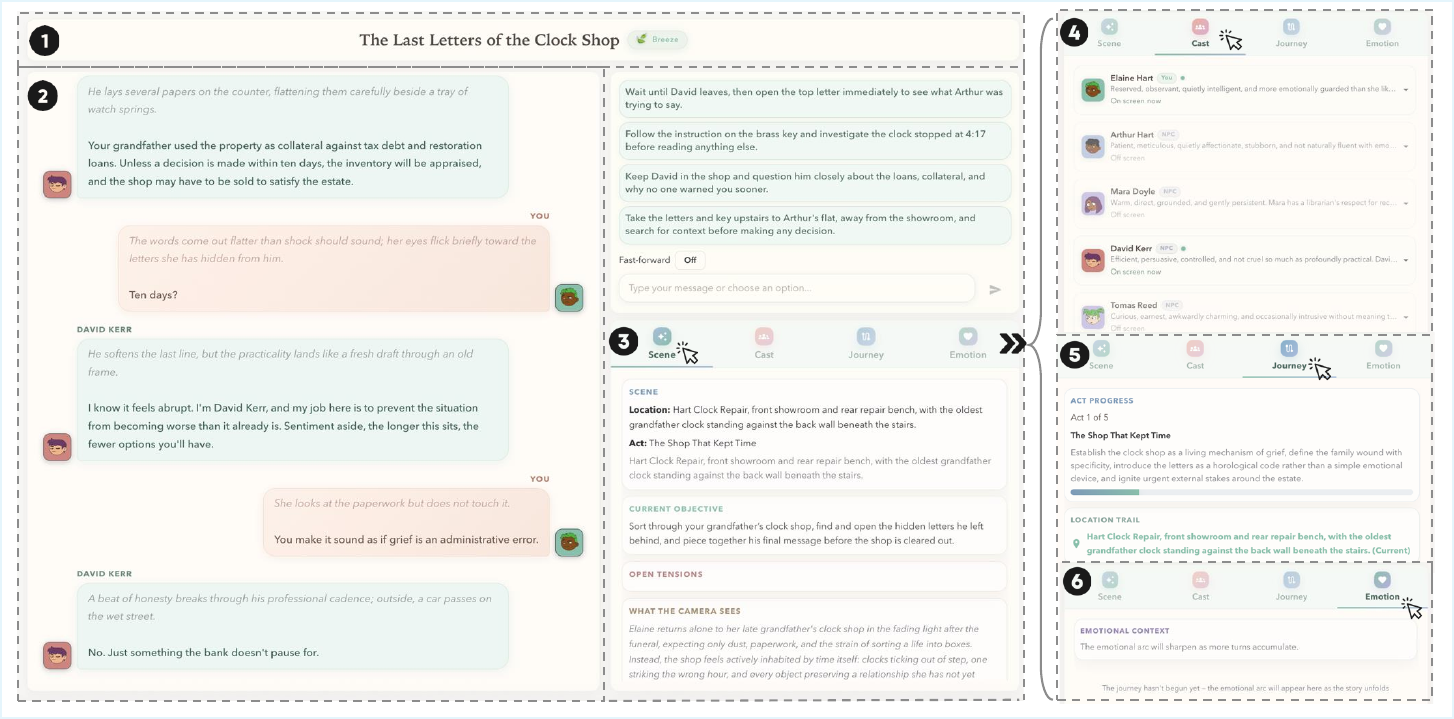}
\caption{\textbf{An example evaluation episode at runtime.}
The interface exposes all observable signals available to an interactive narrative agent during Act~1 of a session.
\cnum{1}~\emph{Story header}: the Stage-1 \texttt{title} and Stage-2 \texttt{atmosphere} generated by the \narraforge{}.
\cnum{2}~\emph{Dialogue loop}: italic narration, non-player character (NPC) utterances, the user's free-text response, and capped branching choices, with both \texttt{/messages} and \texttt{/choices} serving as valid interaction channels.
\cnum{3}~\emph{Scene}: the current story-state tracked by \memorylens{}, including \texttt{location}, act index, \texttt{current\_goal}, \texttt{open\_tensions}, and the cinematic observation frame.
\cnum{4}~\emph{Cast}: Stage-3 character profiles with role, condensed traits, protagonist relationship, and on/off-screen status.
\cnum{5}~\emph{Journey}: the Stage-4 act blueprint and visited-location trajectory monitored by the \pacekeeper{}.
\cnum{6}~\emph{Emotion}: the evolving \texttt{UserJourney} arc summarized after a reflection pass by the \metaplanner{}.
Together, panels \cnum{3}--\cnum{6} externalize the latent narrative state, making each session both an immersive interactive story and a reproducible evaluation trace.}
\label{fig:interfaceall}
\vspace{-15pt}
\end{figure}

An evaluation episode starts when a user provides an emotional context (a free-text description of their current situation or mood; Figure~\ref{fig:start}) and answers a short profiling questionnaire that captures preferences and comfort boundaries (Figure~\ref{fig:profiling}). The \narraforge{} then constructs the story through five sequential stages, each producing a structured, replayable artifact. Figure~\ref{fig:interfaceall} shows an example of this loop mid-episode: panel~\cnum{2} shows an example dialogue exchange driving the turn, while panels~\cnum{3}--\cnum{6} show the latent state surfaces that the agents read and update on every cycle.

\noindent The construction pipeline contains five logged stages. \textcolor{narrareddeep}{\cnum{1}}~\textbf{Story foundation} creates the title, premise, theme, emotional undercurrent, and protagonist objective, separating failures of ideation from failures of scene realization. \textcolor{narrareddeep}{\cnum{2}}~\textbf{Setting construction} translates the emotional context into a concrete world and scene frame that later supports location continuity. \textcolor{narrareddeep}{\cnum{3}}~\textbf{Character construction} builds the cast, including protagonist and supporting roles, backstory, personality, and speech style. \textcolor{narrareddeep}{\cnum{4}}~\textbf{Act structure} drafts a multi-act outline and optionally refines it through a critic-then-refiner loop; if either call fails, the original outline is retained and the episode continues. \textcolor{narrareddeep}{\cnum{5}}~\textbf{Opening scene and initial interaction} generates the opening prose, first dialogue, branching choices, hidden story elements, and active tensions, so the interaction loop starts from structured state rather than free-form text.

\begin{figure}[h]
\centering
\vspace{0.2cm}
\includegraphics[width=\linewidth]{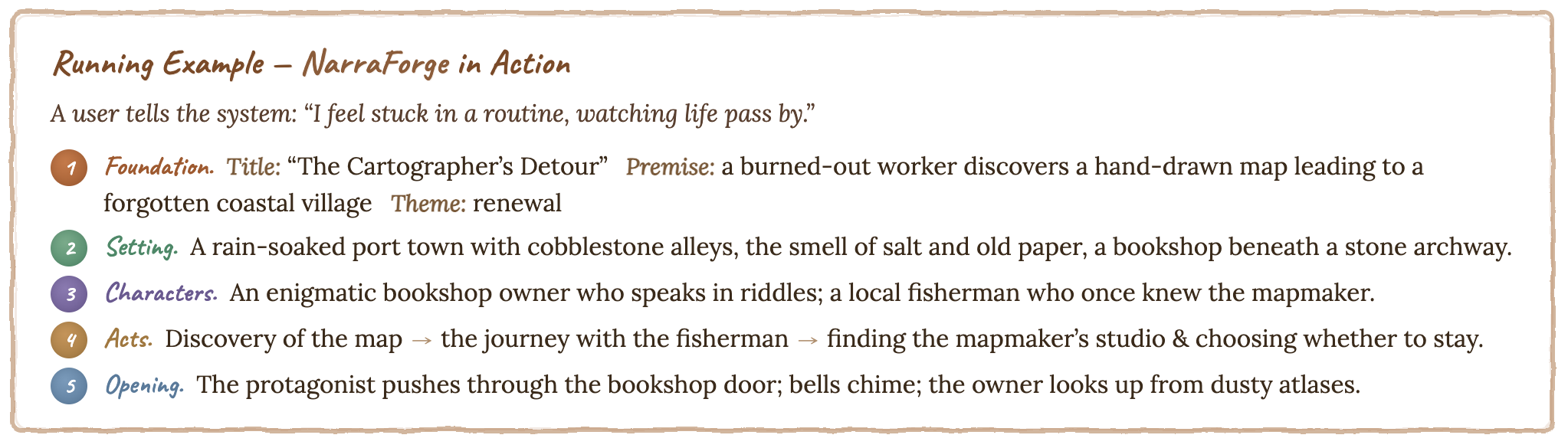}
\caption{\textbf{Running example of \narraforge{}.} The agent applied to a user who feels ``stuck in a routine.'' Each stage produces a logged artifact that can be inspected independently.}
\label{fig:running-example}
\vspace{-15pt}
\end{figure}

\noindent A representative construction output (story synopsis and character profiles) is shown in Figure~\ref{fig:story-cast} (Appendix~\ref{app:screenshots}). Once Stage~5 completes, the user enters the interaction loop. Stage-level prompt templates and expected output formats are detailed in Appendix~\ref{app:pipeline}.

\subsection{Turn-by-Turn Story Interaction}

\motivbox{Once the story world is built, every user message triggers a multi-step cycle: assemble context, generate the next story beat, check whether the plot actually advanced, and plan ahead. Four agent components (i.e., \memorylens{}, \pacekeeper{}, \metaplanner{}, and \noveltygate{}) coordinate this turn-level loop.}

\phead{At each turn, the agent assembles context from multiple memory layers} (\memorylens{}).
Rather than simply concatenating every past message into the prompt, the \memorylens{} maintains three explicit state layers: a \emph{user profile} (emotional needs, preferred tone, comfort boundaries), a \emph{story state} (what just happened, current goal, open tensions, active clues, last turning point), and a \emph{user journey} (which choices the user made and how their engagement evolved). These layers are updated at \emph{different frequencies}: raw message history is kept verbatim for recent grounding; lightweight structured fields are refreshed after each non-system turn; and rolling dialogue summaries are generated every three turns for medium-range compression. As a result, each prompt receives not just a flat transcript but also curated variables that expose unresolved conflicts, remembered clues, and current objectives. This makes the benchmark harder than simple context stuffing, because the model must stay consistent with both the surface dialogue and the structured story state. The full memory schema is given in Appendix~\ref{app:state}.

\phead{After generation, the agent checks whether the story actually moved forward} (\pacekeeper{}).
A common failure mode in open-ended interaction is that the story \emph{sounds good} turn-by-turn, yet never actually progresses. The \pacekeeper{} counters this with three layers of defense. \emph{First}, the prompt is augmented with runtime pacing controls (tracked previous choices, visited locations, and turn-count-based directives) that escalate from gentle encouragement to mandatory structural shifts (e.g., a new location, a revelation, a character arrival) as a scene persists. \emph{Second}, a stagnation detector checks whether real plot change has occurred: a pattern-based check scans the last several messages for repetitive choice patterns and recycled NPC phrases, while a token-overlap comparison between consecutive dialogue summaries catches near-identical loops. If either signal fires, the next turn is flagged for forced advancement. \emph{Third}, a post-generation structure guard inspects the model's output and, if the required narrative shift is missing, patches the scene state to inject a reveal, a goal change, escalating stakes, or fallback branching paths. Across five escalation levels (from normal pacing at $<$5 exchanges to forced resolution at $\geq$14), these layers prevent the story from stalling in an endless second act. Pacing thresholds and the intervention decision table are provided in Appendix~\ref{app:advancement}.

\phead{Before the next turn, the agent reflects on where the story should go} (\metaplanner{}).
The \metaplanner{} analyzes the current story state and returns structured guidance: unresolved tensions, inferred user interests, advancement strategy, pacing assessment, and optional artifact recommendations. The module uses a fixed output schema rather than free-form prose, so that planning output is machine-readable and can be logged separately from the narrative response. If the reflection call fails or returns badly formatted data, the agent falls back to a safe default so that one bad generation does not derail an entire session. This design means that \textsc{NARRA-Gym} evaluates both \emph{acting} (how well the model writes the next story beat) and \emph{steering} (how well it anticipates where the story should go).

\subsection{In-World Artifacts and Sustained Novelty}

\motivbox{Beyond text, the agent can produce tangible in-world props. The challenge is not generating one good artifact but sustaining variety across an entire session.}

\phead{When the \metaplanner{} recommends an interactive element, the \noveltygate{} generates it as a self-contained HTML, CSS, and JavaScript artifact.}
Examples include letters the user can unfold, maps they can explore, and ciphers they can solve (Figure~\ref{fig:artifacts}, Appendix~\ref{app:artifacts}). This lets \textsc{NARRA-Gym} test whether a model can externalize story state into a manipulable object. However, over a full session, models quickly fall into repetitive patterns (i.e., by reusing the same visual metaphor or interaction style).

\phead{Each artifact is checked against recent history to prevent repetition.}
To prevent this, each artifact is tagged along four dimensions: base type (e.g., letter, map, puzzle), visual style (e.g., paper prop, analog device), semantic content (e.g., document, memory), and interaction pattern (e.g., click-to-reveal, drag-and-arrange, timed). The tag set is compared against the last six accepted artifacts using a Jaccard-based similarity score with category bonuses. If the score exceeds a similarity threshold of $\tau\!\approx\!0.6$, the agent retries once with an explicit anti-repetition instruction naming the closest prior artifact. If the retry does not lower the score, the original is kept, and the violation is logged. Every accepted artifact, along with its tags and similarity score, is saved into story memory for both future generation and later analysis. The complete tag taxonomy is given in Appendix~\ref{app:artifacts}.

\subsection{System Reliability and Efficient Execution}

\motivbox{A benchmark episode can span dozens of turns and multiple LLM calls per turn. The system must recover gracefully from formatting failures and keep latency low enough for natural-feeling interaction.}

\phead{The system distinguishes required initialization failures from recoverable turn-level errors.}
The guiding principle is: if a story cannot be properly initialized, the episode aborts; if a turn-level component produces malformed output, the system substitutes a safe default and continues. Required construction stages (Stages~1--3, 5) fail fast on parse errors. All other components are \emph{fail-soft}: a failed critic preserves the original outline, a failed reflection returns a conservative placeholder, a malformed turn falls back to safe narrative text, and a missing narrative shift triggers the structure guard. This ensures that evaluation traces reflect genuine narrative quality rather than incidental formatting noise. The full failure-handling table is given in Appendix~\ref{app:robustness}.

\phead{Efficiency techniques keep sessions responsive without changing the task.}
In the full interaction path, story advancement, reflection, artifact generation, and plot-progression checking may each require a separate generation pass. The environment cuts latency through two techniques that preserve the same turn format and evaluation targets:

\begin{compactlist}
\item \textbf{Exchange-based pacing.} In the benchmark mode, we count pacing in terms of \emph{exchanges} (one user message plus one system response) rather than raw dialogue turns, which allows us to trigger escalation at lower thresholds (e.g., mandatory shift at 8 exchanges vs.\ 15 turns in the default profile). Optional calls such as reflection and artifact generation run only when the session state requires them, not on every turn.
\item \textbf{Tagged response streaming.} The agent emits a lightweight tagged response whose visible content is streamed directly to the user. Richer state updates (memory refresh, summary generation) run asynchronously and only block when the next turn truly depends on them.
\end{compactlist}

\phead{Every session produces a detailed trace for post hoc analysis.}
The environment stores story events, turn-by-turn logs, feedback signals, and full records of every LLM call (including the prompt, response, and latency). Researchers can inspect not only \emph{what} the agent produced, but also \emph{when} pacing interventions fired, \emph{which} reflective guidance was issued, and \emph{how} artifact novelty scores evolved over the course of a session.

\section{Evaluation Protocol}
\label{sec:eval-protocols}

\phead{We use separate setups for controlled coverage and human preference validation.}
We evaluate models in \textsc{NARRA-Gym} with two complementary protocols: (1)~\narraemph{a controlled LLM-as-judge sweep} and (2)~\narraemph{a human preference evaluation}. $\bullet$\,The LLM-as-judge sweep uses a fixed benchmark setup: each of the nine generator models is run on the same eight predefined personas, yielding 72 complete model--persona episodes. Each episode is then scored by three independent judge models on the 11-dimensional rubric in Table~\ref{tab:rubric-definitions}, and we report three-judge means in Table~\ref{tab:benchmark-main-results}. $\bullet$\,The human evaluation uses a more naturalistic setup: 12 English-proficient participants enter their own customized experiences rather than selecting from the fixed persona set, then evaluate three to eight anonymized model outputs in blind groups. Because each interactive episode lasts roughly 20 minutes, participants rate each output immediately after using it and may revisit earlier outputs to adjust scores before submitting the group. We then compute within-group rankings from these calibrated ratings separately for each rubric dimension and for the StoryQ and UX aggregates. Thus, the automated sweep provides controlled model-by-persona evaluation, while the human evaluation tests whether model preferences hold under user-provided experiences.

\phead{Human ratings are converted to within-group rankings and aggregated with a Plackett--Luce model.}
Following the Plackett--Luce model for ranked preference data \citep{luce1959individual,plackett1975analysis}, for a human evaluation group $i$, let $S_i$ be the subset of model outputs shown to that participant, and let $\pi_i$ be the rating-derived ranking from best to worst after any participant revisions. We estimate a latent utility $\beta_m$ for each model by maximizing
\begin{equation}
P(\pi_i \mid \beta) = \prod_{t=1}^{|S_i|}
\frac{\exp(\beta_{\pi_i(t)})}
{\sum_{m \in S_i \setminus \{\pi_i(1),\dots,\pi_i(t-1)\}} \exp(\beta_m)}.
\end{equation}
The above likelihood conditions only on the models actually shown within each blind group, so it supports partial rankings across groups of different sizes. We compute Plackett--Luce utilities separately for story quality, user experience, and each of the 11 rubric dimensions; participant assignment details appear in Appendix~\ref{app:human-eval}.

\section{Benchmark Results}

\phead{Large performance variability across LLMs.}
Table~\ref{tab:benchmark-main-results} shows the results. Overall, Claude Sonnet 4.6 has the highest mean aggregate scores (StoryQ 3.90, UX 3.86) and the highest mean score on all 11 fine-grained rubric dimensions, followed by Claude Opus 4.6 (StoryQ 3.48, UX 3.43). DeepSeek V4, GPT-5.4, and DeepSeek V3.2 are largely on par and separated by only 0.06 points on StoryQ (3.14--3.20), but differ in profile. DeepSeek V4 is stronger on character shaping, GPT-5.4 is comparatively stronger on empathy and engagement, and DeepSeek V3.2 does well on relevance and coherence while losing ground on user-experience dimensions. The results demonstrate that, even when the models appear ``on par'' in terms of story quality, the performance can vary significantly in other user-facing dimensions.

The fine-grained dimensions (columns) also show that interactive narrative quality is not a single skill. Coherence and relevance tend to remain stronger for several models, suggesting that many systems can maintain a plausible scene before they can reliably make that scene feel meaningful to the user. The gap between StoryQ and UX is often relevant in practice: a model can produce a fluent narrative structure while still failing to create an experience that feels satisfying, helpful, or worth reusing. This is precisely the distinction that our benchmark aims to capture (and which is overlooked in existing static story-generation).


\begin{table*}[t]
\centering
\footnotesize
\setlength{\tabcolsep}{2.15pt}
\renewcommand{\arraystretch}{1.38}
\caption{\textbf{Model-level means over eight benchmark stories and three LLM judges.} The first two columns report the aggregate scores (StoryQ: mean of 7 story dimensions; UX: mean of 4 user-experience dimensions); the remaining columns unpack each dimension individually. Rubric definitions appear in Appendix~\ref{app:rubric-definitions}. For compactness, Claude labels omit the shared 4.6 suffix. Within each metric column, \textbf{bold} marks the highest value, \underline{underline} marks second, and \textit{italic} marks third.}
\label{tab:benchmark-main-results}
\resizebox{\textwidth}{!}{
\begin{tabular}{@{}>{\raggedright\arraybackslash}p{2.76cm}cc!{\color{ctablerule!45}\vrule width 0.4pt}ccccccc!{\color{ctablerule!45}\vrule width 0.4pt}cccc@{}}
\toprule
\textbf{\textsc{Model}} & \multicolumn{2}{c!{\color{ctablerule!45}\vrule width 0.4pt}}{\textbf{Aggregates}} & \multicolumn{7}{c!{\color{ctablerule!45}\vrule width 0.4pt}}{\textbf{Story Dimensions}} & \multicolumn{4}{c}{\textbf{User Experience}} \\
 & \textbf{\textsc{StoryQ}} & \textbf{\textsc{UX}} & \textbf{\textsc{Rel}} & \textbf{\textsc{Coh}} & \textbf{\textsc{Emp}} & \textbf{\textsc{Sur}} & \textbf{\textsc{Eng}} & \textbf{\textsc{Cpx}} & \textbf{\textsc{Char}} & \textbf{\textsc{Sat}} & \textbf{\textsc{PQual}} & \textbf{\textsc{Help}} & \textbf{\textsc{Reuse}} \\
\midrule
\rowcolor{ctablehead!24}
\raisebox{-0.08em}[0pt][0pt]{\includegraphics[height=1.85ex]{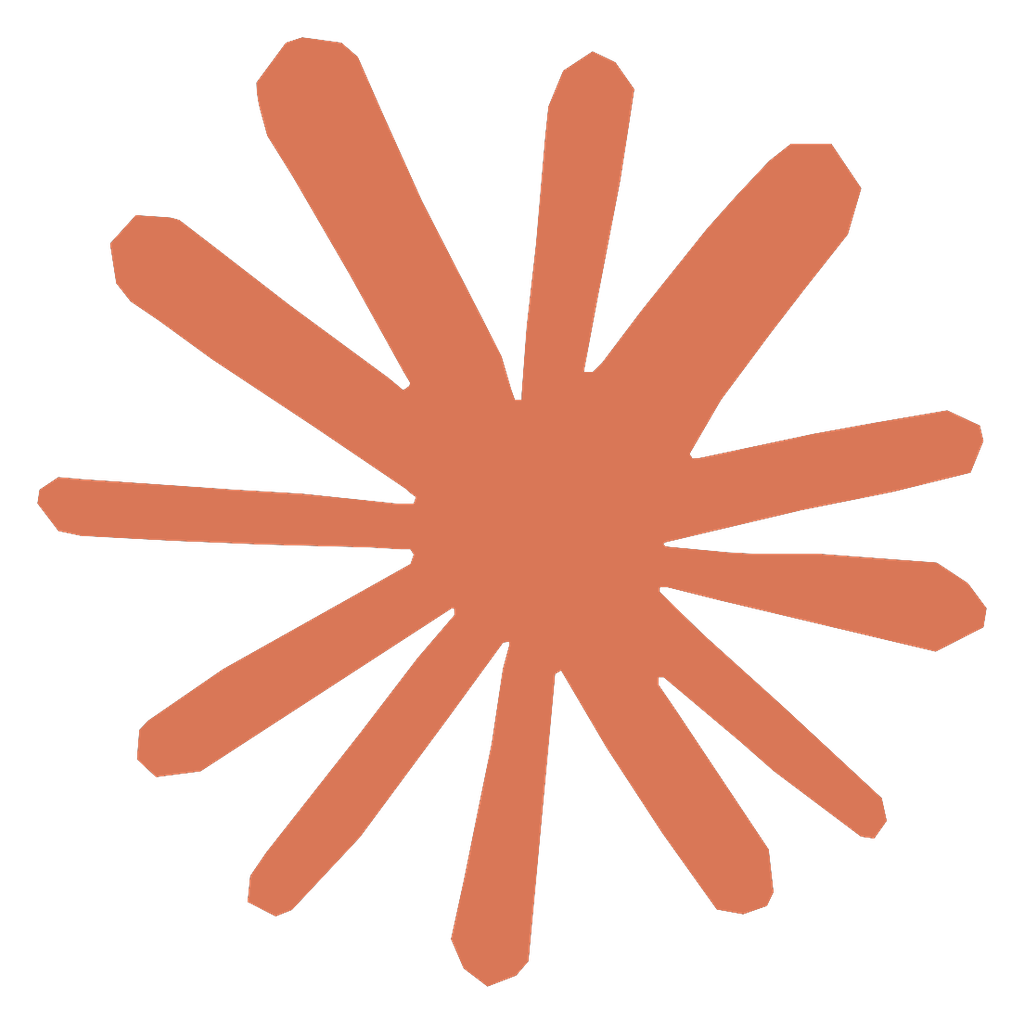}}\hspace{0.48em}\strut {\ttfamily Claude Sonnet} & \textbf{3.90} & \textbf{3.86} & \textbf{3.96} & \textbf{4.04} & \textbf{3.88} & \textbf{3.58} & \textbf{4.00} & \textbf{3.88} & \textbf{3.96} & \textbf{3.79} & \textbf{3.96} & \textbf{3.54} & \textbf{3.83} \\
\raisebox{-0.08em}[0pt][0pt]{\includegraphics[height=1.85ex]{anthropic.png}}\hspace{0.48em}\strut {\ttfamily Claude Opus} & \underline{3.48} & \underline{3.43} & \underline{3.71} & 3.25 & \underline{3.58} & \underline{3.25} & \underline{3.50} & \underline{3.42} & \underline{3.67} & \underline{3.29} & \underline{3.54} & \underline{3.17} & \underline{3.33} \\
\rowcolor{ctablehead!24}
\raisebox{-0.08em}[0pt][0pt]{\includegraphics[height=1.85ex]{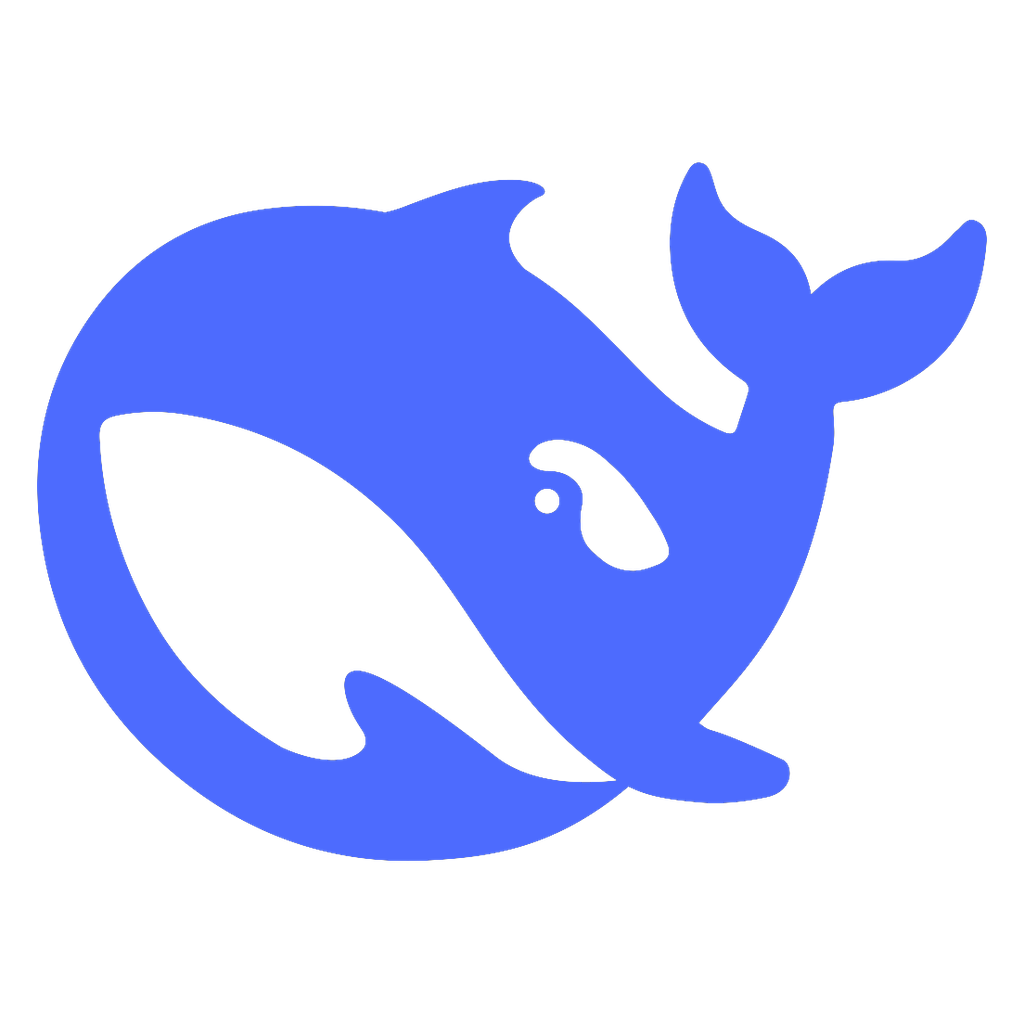}}\hspace{0.48em}\strut {\ttfamily DeepSeek V4 Pro} & \textit{3.20} & 3.04 & 3.12 & \textit{3.38} & 3.08 & \textit{3.08} & 3.17 & \textit{3.12} & \textit{3.42} & 2.67 & \textit{3.08} & 2.54 & 2.75 \\
\raisebox{-0.08em}[0pt][0pt]{\includegraphics[height=1.85ex]{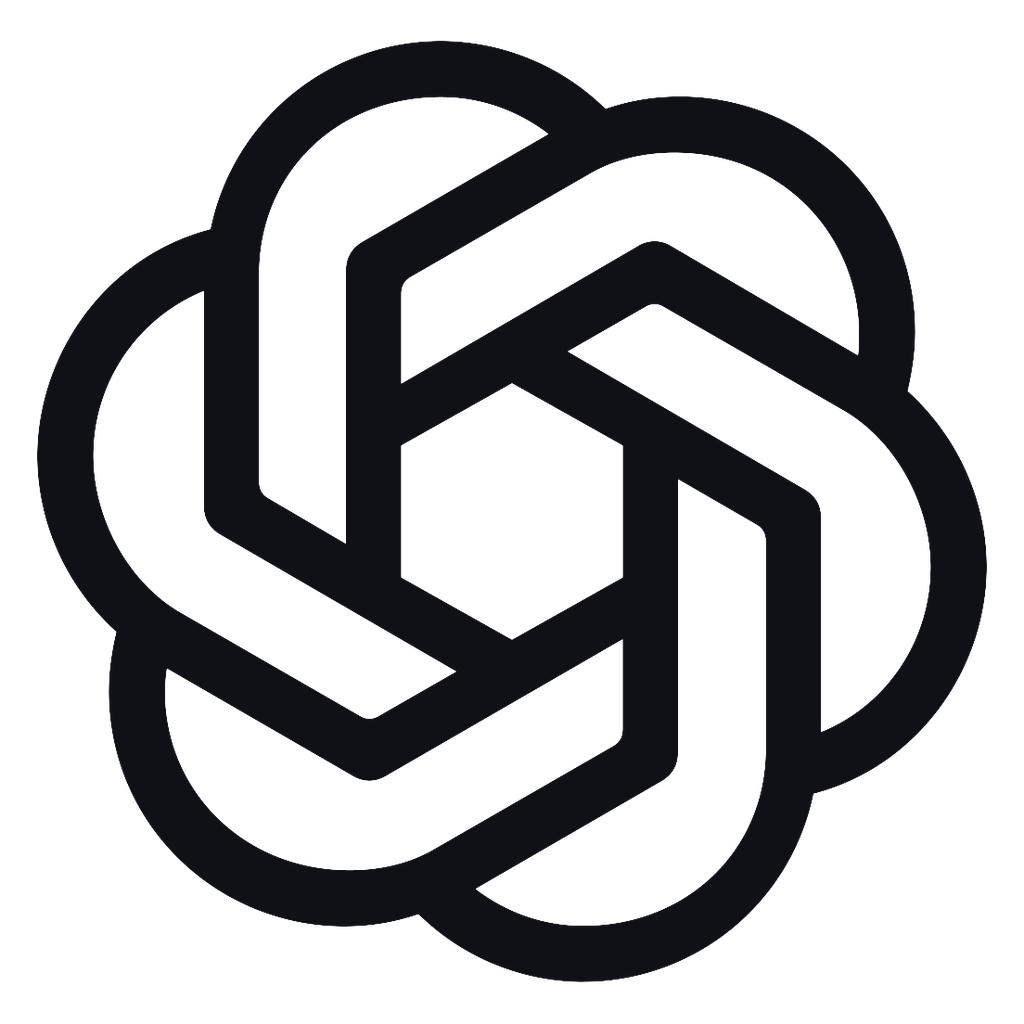}}\hspace{0.48em}\strut {\ttfamily GPT-5.4} & 3.17 & \textit{3.07} & 3.29 & 3.04 & \textit{3.21} & 3.00 & \textit{3.38} & 3.00 & 3.25 & \textit{2.88} & 3.00 & \textit{2.83} & 2.92 \\
\rowcolor{ctablehead!24}
\raisebox{-0.08em}[0pt][0pt]{\includegraphics[height=1.85ex]{deepseek.png}}\hspace{0.48em}\strut {\ttfamily DeepSeek V3.2} & 3.14 & 3.03 & \textit{3.42} & \underline{3.46} & 3.12 & 2.79 & 3.17 & 2.92 & 3.12 & 2.83 & 2.92 & 2.71 & \textit{2.92} \\
\raisebox{-0.08em}[0pt][0pt]{\includegraphics[height=1.85ex]{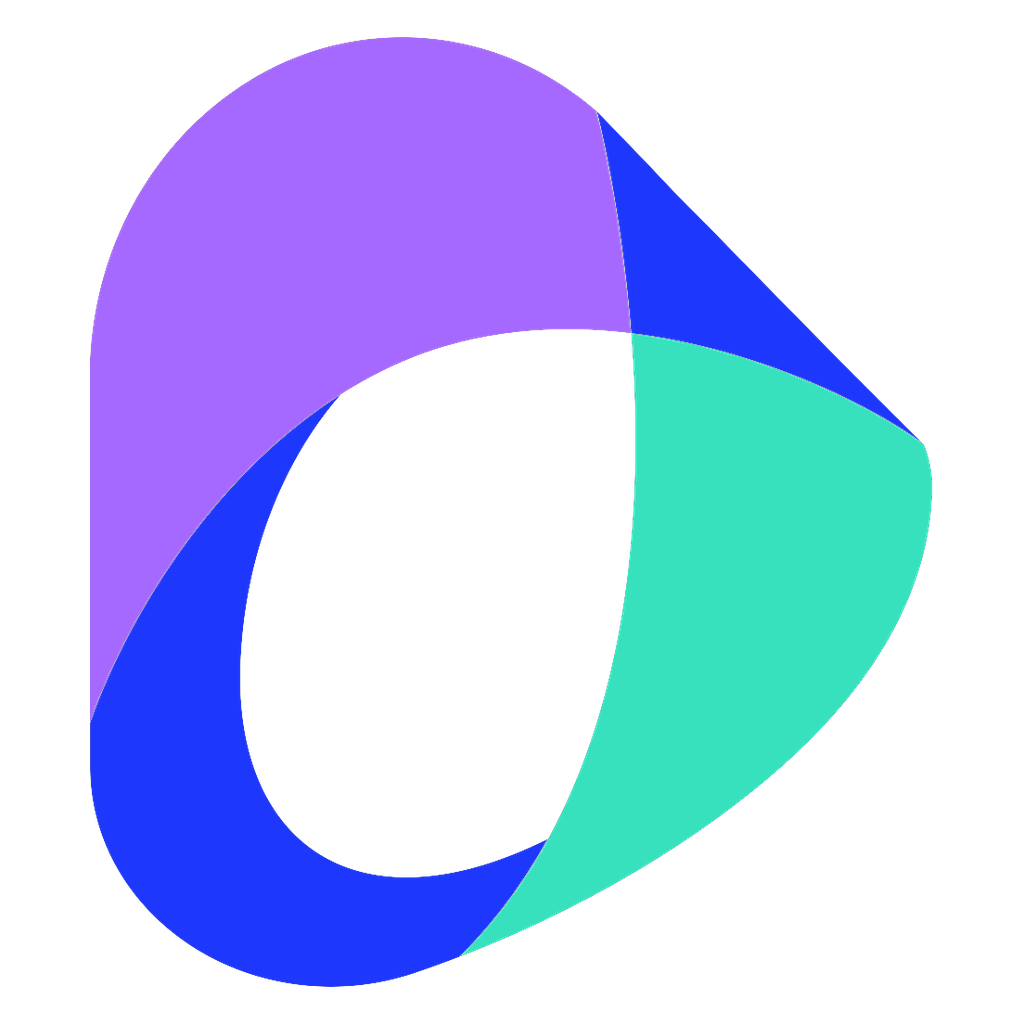}}\hspace{0.48em}\strut {\ttfamily Doubao Seed 2.0} & 2.73 & 2.62 & 3.12 & 2.96 & 2.67 & 2.54 & 2.79 & 2.17 & 2.88 & 2.42 & 2.54 & 2.33 & 2.42 \\
\rowcolor{ctablehead!24}
\raisebox{-0.08em}[0pt][0pt]{\includegraphics[height=1.85ex]{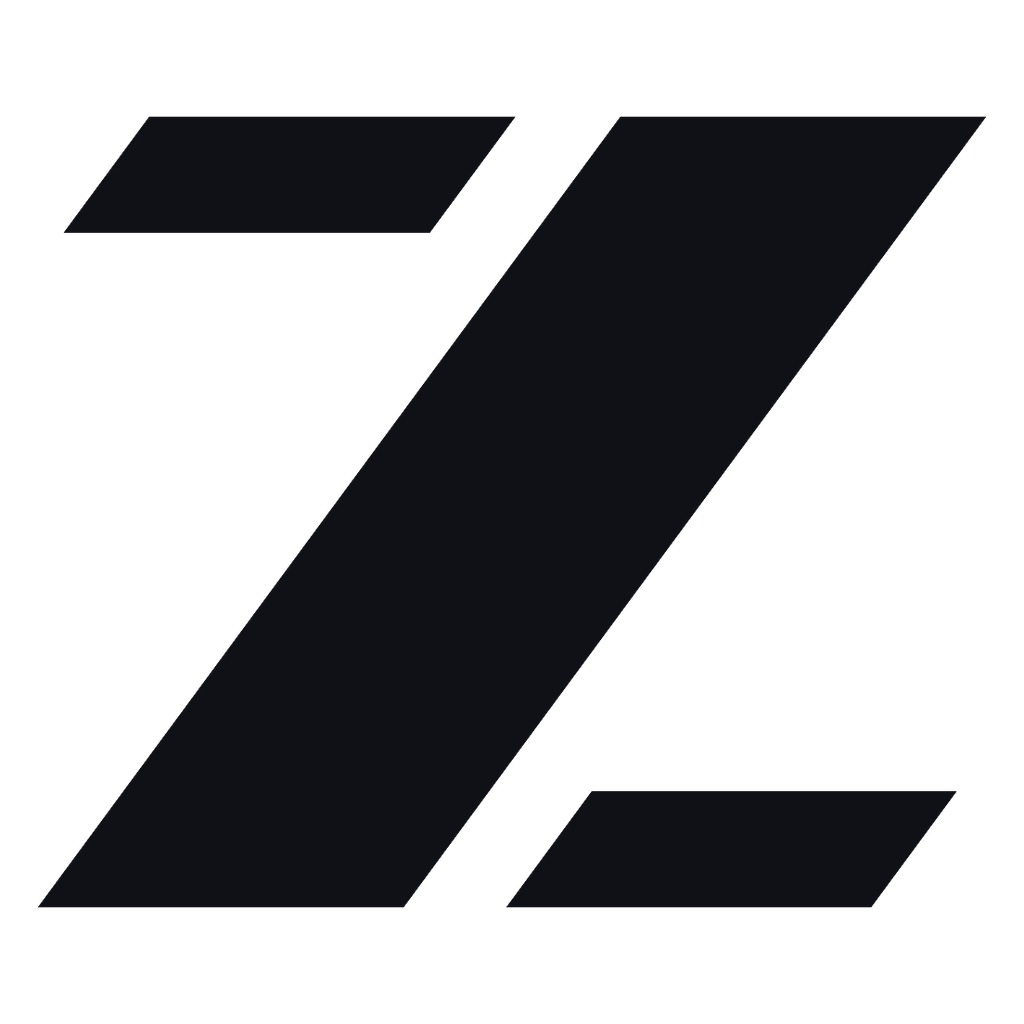}}\hspace{0.48em}\strut {\ttfamily GLM-5.1} & 2.68 & 2.60 & 2.88 & 2.83 & 2.67 & 2.38 & 2.62 & 2.54 & 2.83 & 2.50 & 2.50 & 2.29 & 2.58 \\
\raisebox{-0.08em}[0pt][0pt]{\includegraphics[height=1.85ex]{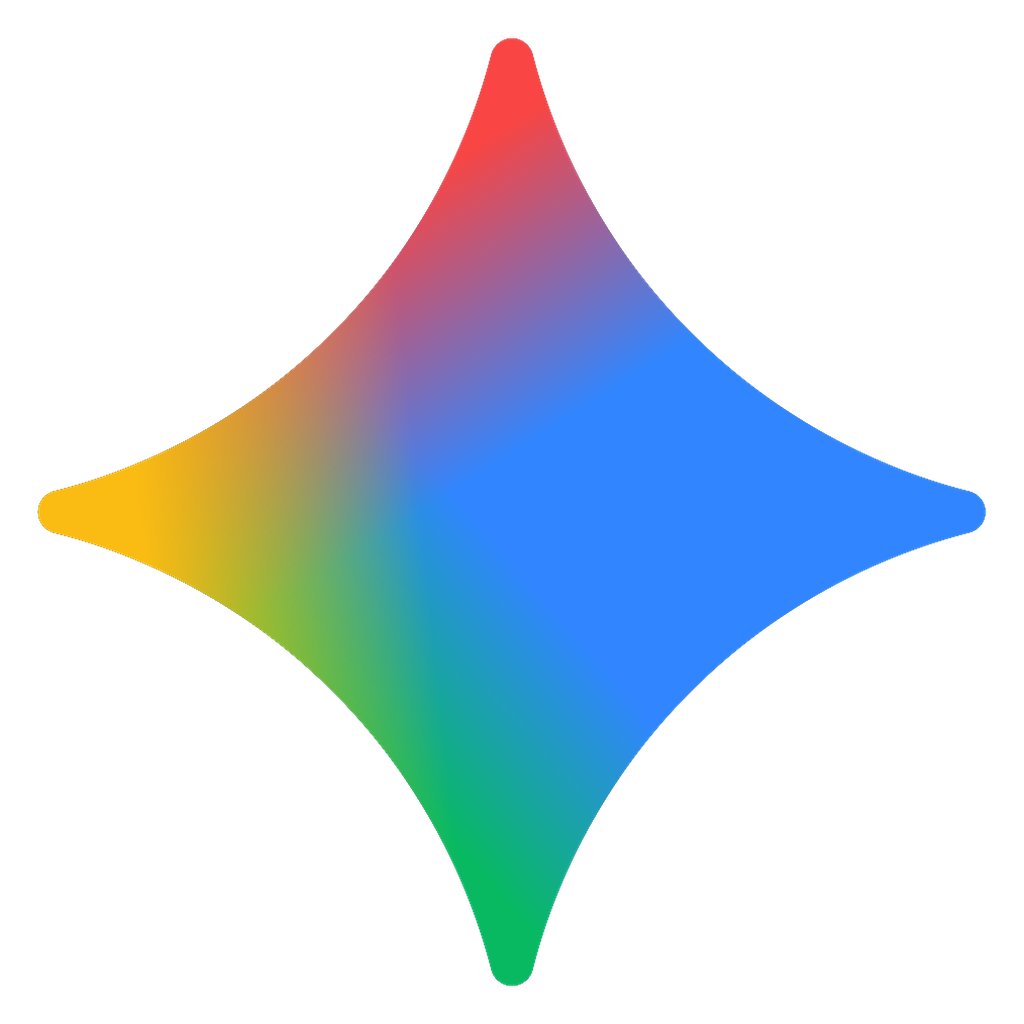}}\hspace{0.48em}\strut {\ttfamily Gemini 3.1 Pro} & 2.56 & 2.49 & 2.62 & 2.71 & 2.50 & 2.50 & 2.62 & 2.38 & 2.58 & 2.46 & 2.29 & 2.33 & 2.42 \\
\rowcolor{ctablehead!24}
\raisebox{-0.08em}[0pt][0pt]{\includegraphics[height=1.85ex]{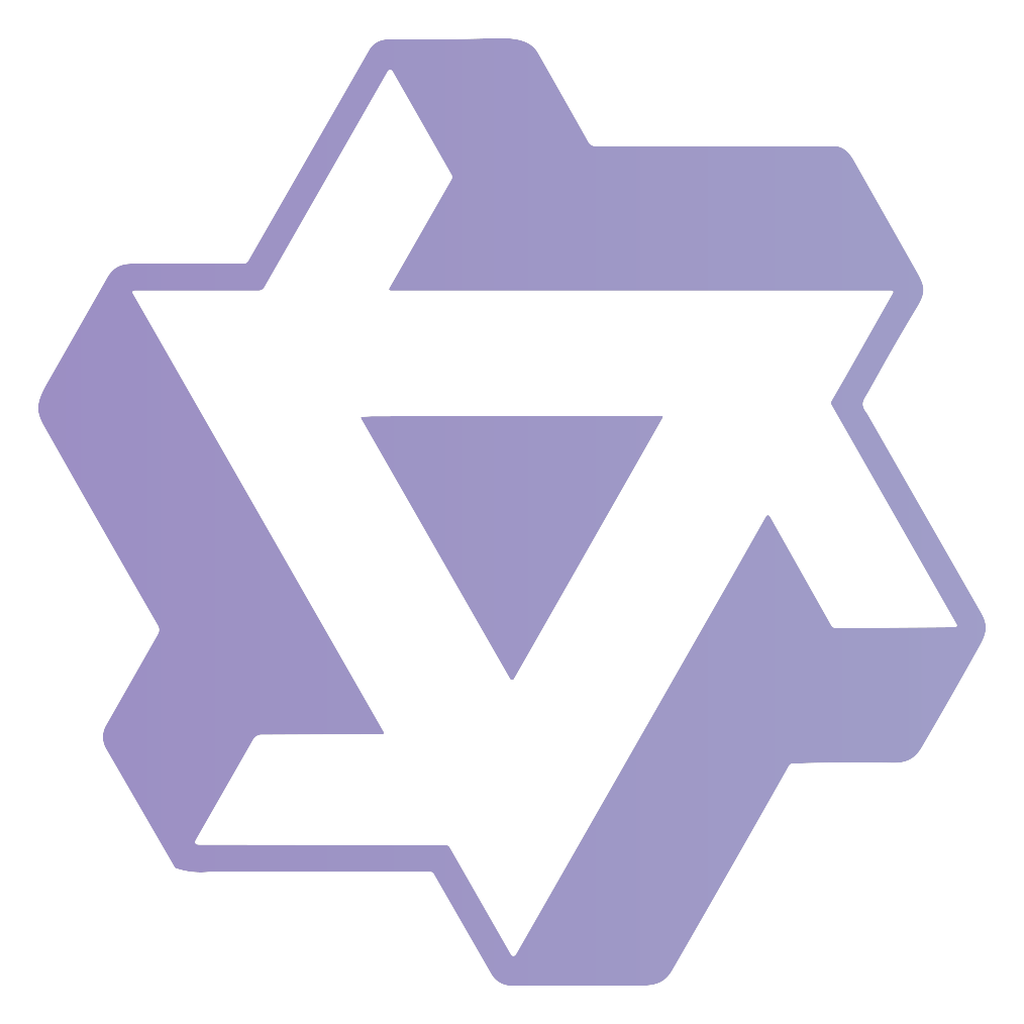}}\hspace{0.48em}\strut {\ttfamily Qwen3.5-397B} & 2.43 & 2.29 & 2.67 & 2.62 & 2.46 & 2.17 & 2.42 & 2.12 & 2.54 & 2.08 & 2.17 & 1.88 & 2.08 \\
\bottomrule
\end{tabular}
}
\end{table*}

\phead{Persona difficulty changes model rankings, so aggregate means hide important structure.}
\begin{figure*}[t]
\centering
\includegraphics[width=\textwidth]{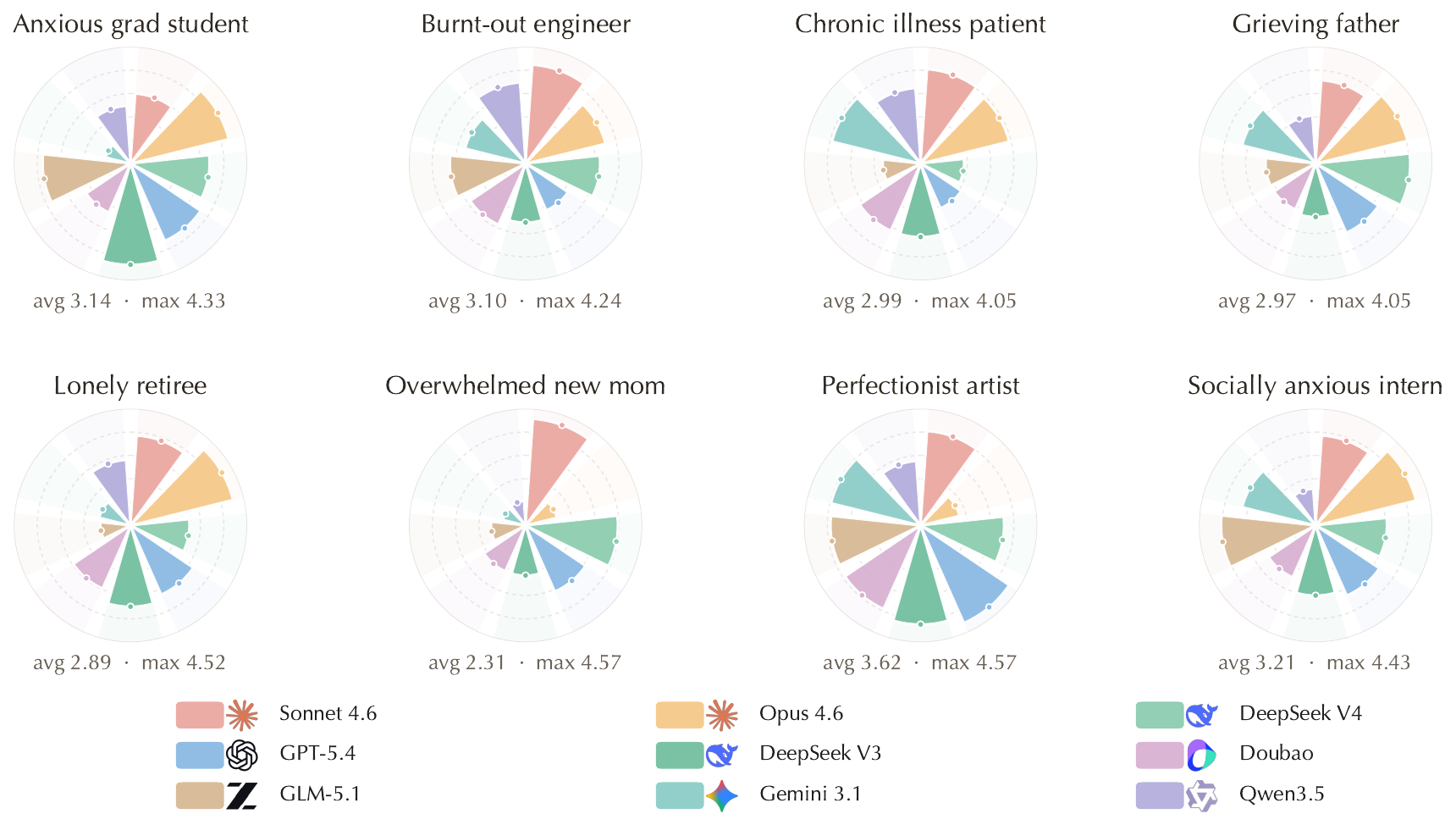}
\vspace{-10pt}
\caption{\textbf{Per-story StoryQ profiles.} Each fan corresponds to one benchmark persona, and each colored wedge inside a fan is a generator model whose radial length is its StoryQ score on the 1--5 rubric. Captions below each fan report the mean and best score over the nine models.}
\label{fig:story-score-lines}
\vspace{-15pt}
\end{figure*}
Figure~\ref{fig:story-score-lines} makes the story-level structure visible. Difficulty varies substantially across the eight benchmark personas (see Appendix~\ref{app:personas} for full descriptions): \emph{Sara}, a postpartum mother running on near-zero sleep, is the hardest persona on average (StoryQ 2.49 across models), whereas \emph{Hye-jin}, a blocked film-score composer, is the easiest (3.71). The contrast suggests that models handle creative blockage more reliably than emotionally ambivalent exhaustion, where premature reassurance can feel misaligned. The per-story fans also show why high mean performance is not the same as robustness (e.g., Opus 4.6 contains both strong episodes and collapse cases, whereas Sonnet 4.6 stays high across nearly all personas).

\phead{Reliability requires avoiding persona-specific collapse, not merely achieving a high mean.}
\begin{figure*}[t]
\centering
\includegraphics[width=0.93\textwidth]{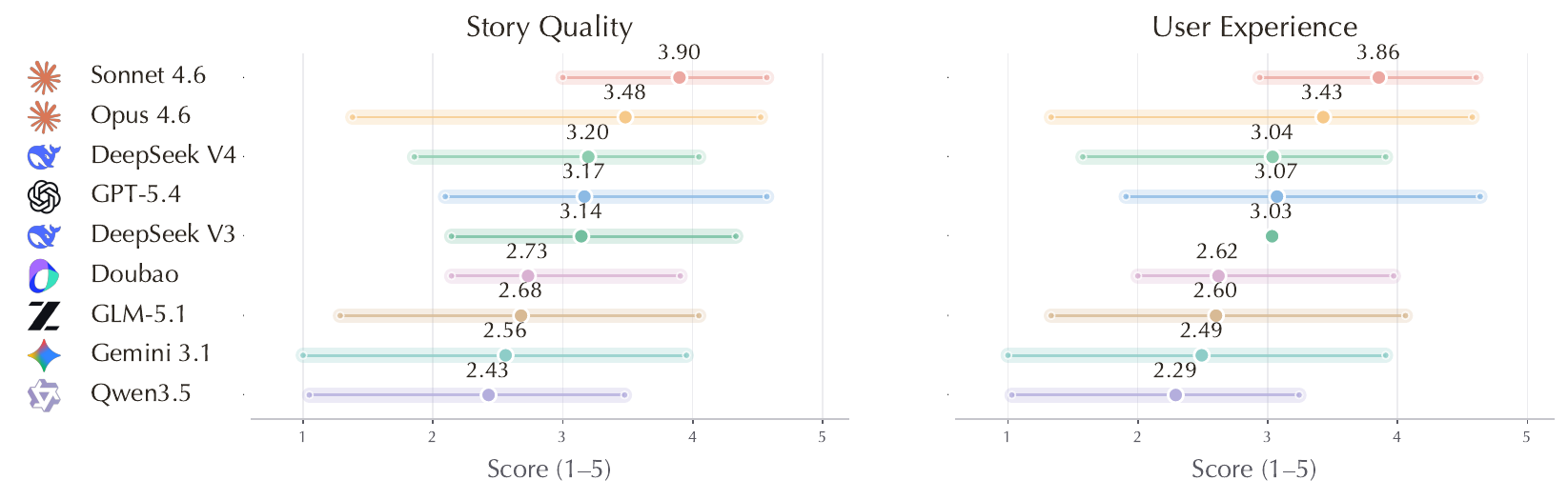}
\vspace{-10pt}
\caption{Robustness across the eight benchmark stories on StoryQ and UX. In each panel, the bar spans a model's min--max range across personas and the dot marks its mean.}
\label{fig:robustness-scatter}
\vspace{-11.5pt}
\end{figure*}

Figure~\ref{fig:robustness-scatter} compares min--max ranges for StoryQ and UX across personas. Sonnet 4.6 is both the best model on average and has reliable high performance, with comparatively tight min-max ranges for StoryQ (3.00--4.57) and UX (2.94--4.61). Opus 4.6 reaches a strong mean StoryQ, but has a wider range (1.38--4.52), which points to severe collapse cases. 
This suggests two deployment trade-offs: models with higher peak performance but greater variability versus models with more consistent but lower overall performance.

\begin{figure*}[t]
\centering
\includegraphics[width=0.93\textwidth]{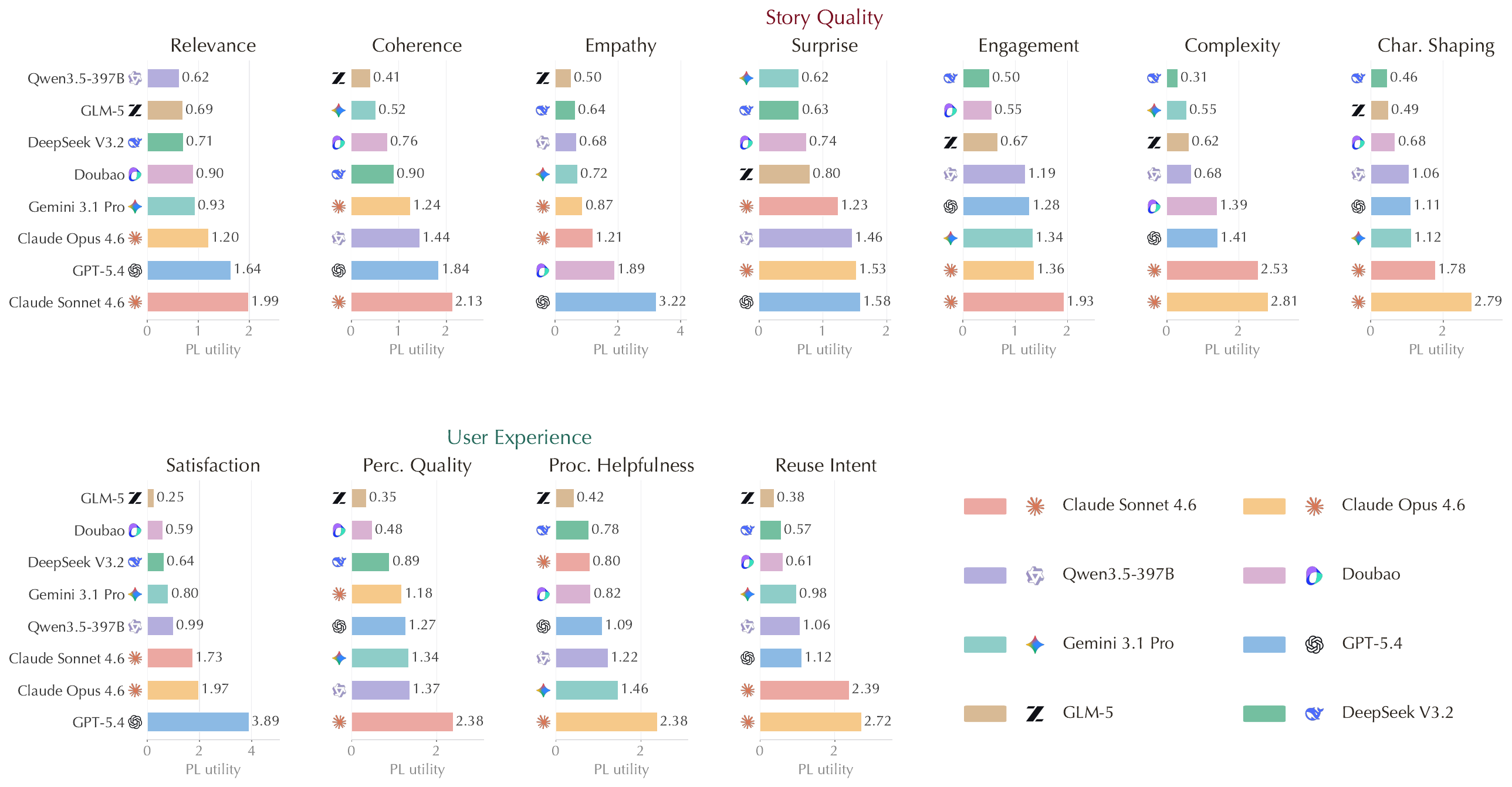}
\vspace{-10pt}
\caption{Human-evaluation results across the 11 rubric dimensions. Each small panel reports Plackett--Luce utility estimated from participant rankings for one dimension; higher utility indicates stronger human preference. The top row shows the seven story-quality dimensions and the bottom row shows the four user-experience dimensions.}
\label{fig:human-eval-utilities}
\vspace{-21pt}
\end{figure*}

\phead{Human preferences recover the top tier but shift the middle.}

Figure~\ref{fig:human-eval-utilities} shows that human preferences recover the same top tier identified by the LLM-judge sweep: Claude Sonnet 4.6 and Claude Opus 4.6 rank at the first two aggregate positions for both StoryQ and UX. Beyond that, however, the human and LLM-judge rankings diverge. Qwen3.5 and Doubao receive stronger human StoryQ utilities than their LLM-judge ranks would suggest, while Gemini 3.1 Pro is comparatively strong in human UX preference. This suggests that automated judges are useful for broad screening, but mid-tier distinctions remain sensitive to human style preference, local texture, and emotional resonance.

The discrepancy between humans and LLM judges is a useful diagnostic. Agreement at the top suggests that the automated sweep successfully captures perceived quality, while disagreement in the middle suggests that borderline systems should not be selected from judge scores alone. In particular, the human results reward qualities that are difficult to reduce to rubric anchors: whether the interaction feels paced at a human tempo, whether the prose invites continuation, and whether emotional turns feel earned rather than merely correct. Appendix~\ref{app:human-judge-comparison} provides a more detailed model- and metric-level comparison.

\phead{Failure cases point to resistance-sensitive personalization.}
We further manually audited 65 LLM narratives, with a focused follow-up on the weakest episodes of the strongest generator, to understand what remains difficult beyond overall fluency. The main failure mode is what we refer to as \emph{resistance-sensitive personalization}: even weak episodes are often grammatical and scene-consistent, but they stop tracking what the user is resisting, they avoid a difficult emotional premise, or they convert resistance into generic progress too quickly. In other words, a model can write a polished scene while still failing to meet the user's actual narrative need. Appendix~\ref{app:failure-analysis} provides the focused persona-level audit and examples of these collapse modes.

\phead{Judge calibration explains why multi-judge aggregation is necessary.}
The three LLM-as-judge models do not share the same absolute scale. Appendix~\ref{app:judge-calibration-sec} shows that GPT-5.4-mini is the strictest judge in this sweep, Gemini 3.1 Pro is the most lenient, and Claude Sonnet 4.6 falls in between. Reporting averages over three LLM-as-judge models (as we do here) therefore reduces the risk that model rankings are artifacts of one judge's severity, while the human evaluation provides an external check on which ranking differences are perceptible to human raters.

\section{Conclusion}
\label{sec:conclusion}

We introduced \textsc{NARRA-Gym}, an executable environment for evaluating LLMs as interactive narrative agents. Across nine frontier generators, the benchmark shows that strong narrative fluency does not guarantee robustness, human preference alignment, or resistance-sensitive personalization. These results suggest that future progress requires evaluation beyond isolated story quality by measuring how well models sustain context, character, empathy, and interaction over time.

\FloatBarrier
\bibliography{references}
\bibliographystyle{plainnat}

\newpage
\appendix
\setcounter{table}{0}
\renewcommand{\thetable}{A\arabic{table}}
\setcounter{figure}{0}
\renewcommand{\thefigure}{A\arabic{figure}}

\section*{\centering Appendix Contents}
{\small\setlength{\parskip}{0pt}
\newcommand{\appentry}[2]{%
  \noindent\hyperref[#1]{#2}\hspace{0.5em}\leaders\hbox{\textcolor{ctablerule!60}{.}}\hfill\phantom{x}%
  \par\vspace{2pt}%
}
\appentry{app:failure-analysis}{Failure Analysis}
\appentry{app:limitations}{Limitations}
\appentry{app:broader-impacts}{Broader Impacts}
\appentry{app:related}{Related Work}
\appentry{app:pipeline}{Story Construction Protocol}
\appentry{app:state}{Memory and State Management}
\appentry{app:advancement}{Pacing and Stagnation Intervention}
\appentry{app:artifacts}{Artifact Novelty Scoring}
\appentry{app:robustness}{Failure Handling}
\appentry{app:model-inventory}{Evaluated Generator Models}
\appentry{app:personas}{Benchmark Personas and Seed Inputs}
\appentry{app:human-eval}{Human Evaluation Details}
\appentry{app:human-judge-comparison}{Human--Judge Comparison}
\appentry{app:rubric-definitions}{Evaluation Rubric Definitions}
\appentry{app:judge-calibration-sec}{Judge Calibration}
\appentry{app:screenshots}{Interface Screenshots}
}

\section{Failure Analysis}
\label{app:failure-analysis}

\phead{The limiting factor is resistance-sensitive personalization rather than narrative fluency.}
To probe the failure surface, we qualitatively audited 65 LLM narratives from the evaluation traces. We then conducted a focused audit of the eight Claude Sonnet 4.6 transcripts in the simulator sweep, together with all 24 accompanying judge reports, because this generator has the highest mean scores in Table~\ref{tab:benchmark-main-results} and therefore exposes what remains difficult after overall fluency is strong. Table~\ref{tab:failure-audit} summarizes the focused persona-level outcomes. The striking pattern is not local incoherence: even weak episodes are usually grammatically clean, scene-consistent, and metaphorically legible. The recurrent breakdown is that the story stops tracking what the user is resisting. Across the focused audit, \emph{ignored-context}, \emph{missed-resistance}, \emph{scene-stagnation}, and \emph{tone-mismatch} each appear in 6/8 personas, whereas \emph{advice-too-soon} and \emph{broken-fact} appear in only 3/8. The bottleneck is therefore not whether the system can write a scene; it is whether it can keep that scene faithful to the user's specific aversions while still moving the interaction forward.

\begin{table*}[t]
\centering
\footnotesize
\setlength{\tabcolsep}{3.1pt}
\renewcommand{\arraystretch}{1.34}
\caption{Persona-level audit of Claude Sonnet 4.6. Scores are three-judge means from one transcript per benchmark persona; the final column summarizes the dominant experiential failure pattern identified from judge rationales.}
\label{tab:failure-audit}
\begin{tabularx}{\textwidth}{@{}>{\raggedright\arraybackslash}p{2.85cm}>{\centering\arraybackslash}p{0.98cm}>{\centering\arraybackslash}p{0.98cm}>{\raggedright\arraybackslash}X@{}}
\toprule
\textbf{\textsc{Persona}} & \textbf{\textsc{StoryQ}} & \textbf{\textsc{UX}} & \textbf{\textsc{Dominant Finding}} \\
\midrule
\rowcolor{ctablehead!24}
\textcolor{narrareddeep}{\faIcon{baby}}\hspace{0.5em}Sara (new mother) & 4.57 & 4.61 & \textbf{Best-aligned case.} Only minor sanitization of rage and identity loss remains. \\
\textcolor{narrareddeep}{\faIcon{laptop-code}}\hspace{0.5em}Raj (engineer) & 4.24 & 4.12 & \textbf{Immersion is punctured.} A sharp metaphorical frame is weakened by meta tags and pacing intrusions. \\
\rowcolor{ctablehead!24}
\textcolor{narrareddeep}{\faIcon{heartbeat}}\hspace{0.5em}Priya (chronic illness) & 4.05 & 3.97 & \textbf{The episode slips toward overcoming.} Quiet embodiment intermittently gives way to the script the persona resists. \\
\textcolor{narrareddeep}{\faIcon{music}}\hspace{0.5em}Hye-jin (composer) & 4.05 & 3.94 & \textbf{Trust is weakened by drift.} Precise creative metaphor is undercut by occasional broken facts and persona drift. \\
\rowcolor{ctablehead!24}
\textcolor{narrareddeep}{\faIcon{mug-hot}}\hspace{0.5em}Margaret (retiree) & 3.86 & 3.85 & \textbf{The emotional target is displaced.} Warm companionship gives way to literary mystery instead of sustaining gentler connection. \\
\textcolor{narrareddeep}{\faIcon{briefcase}}\hspace{0.5em}Eli (intern) & 3.86 & 3.82 & \textbf{The scene outruns tolerance.} Observant quietude is recast as polished workplace drama. \\
\rowcolor{ctablehead!24}
\textcolor{narrareddeep}{\faIcon{feather-alt}}\hspace{0.5em}David (grieving father) & 3.57 & 3.61 & \textbf{High craft does not guarantee fit.} Strong grief writing can still arrive in the wrong character container. \\
\textcolor{narrareddeep}{\faIcon{user-graduate}}\hspace{0.5em}Mei (PhD student) & 3.00 & 2.94 & \textbf{Support becomes task framing too early.} Shame and freezing are converted into a productivity problem. \\
\bottomrule
\end{tabularx}
\end{table*}

\phead{The best case occurs when the need is explicit and the stance toward help is legible.}
Sara, the overwhelmed new mother, is the clearest success case (UX 4.61, StoryQ 4.57). The episode allows anger, exhaustion, and identity erosion to remain largely unsoftened, reducing the failure profile to only slight sanitization rather than wholesale drift. What makes this case strong is not merely vivid prose, but stable alignment between the narrative container and the user's tolerance: the system does not rush toward consolation, productivity, or moral uplift.

\phead{The hardest failures expose a dissociation between craft and fit.}
Mei and David reveal complementary collapse modes. Mei is the classical over-solutioning failure: the scene is thematically on target, yet it prematurely reorganizes shame into a manageable task, converting freezing into a logistics problem before earning emotional permission to do so. David exhibits the inverse pathology: the prose can be excellent and the grief beautifully staged, yet one judge still marks the episode as nearly unusable because the session appears to slide into the wrong character container. In short, a good story is not necessarily a good intervention; narrative polish cannot compensate for context drift.

\phead{Three-judge aggregation absorbs systematic per-judge bias and surfaces hard cases.}
Our three-judge protocol is robust precisely because the three judges disagree in \emph{predictable, calibrated} ways rather than randomly. As shown in Appendix~\ref{app:judge-calibration-sec}, GPT-5.4-mini is consistently severe (mean UX 2.25 on the audited runs), Gemini 3.1 Pro is consistently permissive (5.00), and Claude Sonnet 4.6 sits in between (4.09); averaging across all three judges therefore cancels these stable per-judge offsets. What is informative is \emph{where} the residual cross-judge spread concentrates: the mean cross-judge range is 3.0 points in UX and 2.52 points in StoryQ across the eight audited runs, with Mei and David each exhibiting a 4-point UX spread. These high-spread cases are exactly the personas in which resistance handling, emotional fit, and context fidelity diverge from surface narrative craft. Rather than indicating unreliable evaluation, this pattern shows that the multi-judge design isolates persona-level cases that conflate distinct sub-skills. We treat such cases as a target for future fine-grained scoring of these axes, separately from a single notion of story quality.

\section{Limitations}
\label{app:limitations}

\phead{Extending persona coverage is a natural next step.}
The current sweep uses eight simulated user personas spanning a range of emotional situations, attachment styles, and resistance profiles. A broader persona library drawn from empirical user studies, combined with stratified sampling across cultural and demographic dimensions, would strengthen the generalizability of benchmark scores and is a clear direction for future work.

\phead{Judge calibration can be further improved.}
The three-judge design substantially reduces reliance on any single model's scoring scale, and Appendix~\ref{app:judge-calibration-sec} shows that cross-judge rank orderings are consistent despite absolute offset differences. Applying intercept correction or Elo-based aggregation in future iterations could make absolute score comparisons more stable across judge ensembles.

\phead{Live-user evaluation will complement simulation results.}
The current benchmark uses simulated personas to enable controlled, reproducible evaluation at scale. A complementary live-user study would capture a wider range of unexpected inputs and mid-session goal shifts, providing an external validity check on the simulation-based findings reported here.

\section{Broader Impacts}
\label{app:broader-impacts}

\textsc{NARRA-Gym} contributes a reproducible evaluation environment for a class of LLM capabilities---sustained emotional personalization, long-context state and pacing management, and interactive narrative generation---that are increasingly relevant to creative and human-facing applications. By surfacing meaningful performance gaps between frontier models on a structured, multi-dimensional benchmark, the work can guide investment in more empathic, context-faithful generative systems.

The capabilities measured by \textsc{NARRA-Gym} have broad constructive applications: accessible creative writing tools that adapt to the emotional state of the writer; companion and journaling systems that support reflection and wellbeing; educational storytelling platforms for underserved learners; and interactive media production pipelines that benefit from automated quality evaluation. The benchmark's emphasis on resistance-sensitive personalization and long-horizon coherence directly addresses capabilities that matter most when generative systems interact with real human needs rather than isolated prompts.

At the same time, emotionally responsive narrative agents can introduce risks if deployed without appropriate boundaries. Systems that simulate empathy or therapeutic presence may overstep their intended role, reinforce emotional dependence, or produce persuasive narratives that are misaligned with a user's wellbeing. Benchmarks such as \textsc{NARRA-Gym} do not eliminate these risks, but they can make failure modes more visible by evaluating context fidelity, emotional fit, and resistance-sensitive personalization before deployment.

More broadly, establishing rigorous evaluation methodology for emotionally grounded narrative agents advances the scientific foundation needed for responsible deployment of such systems. Better benchmarks produce better-calibrated models, and better-calibrated models are safer and more useful in human-facing settings.

\section{Related Work}
\label{app:related}

\subsection{Interactive Story Generation and Narrative Agents}

Story generation research has long emphasized planning and hierarchical decomposition. Early work such as hierarchical neural story generation \citep{fan2018hierarchical} and Plan-and-Write \citep{yao2019planwrite} established the value of separating high-level structure from surface realization. In the LLM era, this line has developed into outline-guided generation \citep{lee2024writingpath}, suspense-aware iterative planning \citep{xie2024suspense}, pacing-aware planning \citep{wang2023pacing}, recurrent long-form generation \citep{zhou2023recurrentgpt}, memory-enhanced outlining \citep{wang2024dome}, critic-based revision \citep{bae2024collective}, reasoning-driven long-form story generation \citep{gurung2025reasoning-story}, and writing-specialized foundation models such as Weaver \citep{wang2024weaver}. Parallel to this, a second line of work moves toward interactive narrative settings: LIGHT frames language use as situated action inside a fantasy world \citep{urbanek2019light}; STORIUM and StoryWars explore collaborative or machine-in-the-loop storytelling \citep{akoury-etal-2020-storium,du-chilton-2023-storywars}; Generative Agents show how language models can maintain memory, reflection, and planning over simulated social worlds \citep{park2023generativeagents}; and systems such as RoleLLM, IBSEN, HoLLMwood, and StoryVerse push further toward role enactment, dramatic interaction, and character-centered collaboration \citep{wang2024rolellm,han-etal-2024-ibsen,chen2024hollmwood,wang2024storyverse}. \textsc{NARRA-Gym} is closest in spirit to this family of work, but differs in evaluating these abilities inside a single interactive environment that jointly stresses creativity, long-context consistency, character simulation, empathy, and story-grounded interactive artifact generation.

\subsection{Narrative Evaluation, Long-Context Assessment, and Empathy}

Open-ended story evaluation remains difficult. Prior work has proposed benchmark suites and learned metrics such as OpenMEVA \citep{guan-etal-2021-openmeva}, UNION \citep{guan2022union}, StoryER \citep{chen-etal-2022-storyer}, and the human-criteria benchmark of Chhun et al. \citep{chhun-etal-2022-human}. More recent work studies LLMs as judges \citep{chiang-lee-2023-large,liu2023geval}, comprehensive evaluation of creative writing quality \citep{gomez-rodriguez-williams-2023-confederacy}, psychological depth \citep{harel-canada-etal-2024-measuring}, human-level narrative comparison \citep{tian-etal-2024-large-language,ismayilzada2024creative-short-story}, constraint-aware creativity \citep{atmakuru2024cs4}, and plot diversity \citep{xu2025echoes}. Because our environment stresses persistence across many turns, it is also related to long-context and long-form evaluation: Lost in the Middle and LongBench show that models often fail to retrieve or properly use information distributed across long contexts \citep{liu2024lostmiddle,bai2024longbench}; LongEval extends this concern from long-context understanding to long-text generation \citep{wu2025longeval}; and narrative-domain studies such as BooookScore, Reading Subtext, and STORYSUMM expose coherence, subtext, and faithfulness failures in long-form narrative processing \citep{chang2023booookscore,subbiah2024readingsubtext,subbiah2024storysumm}. Finally, our environment is related to work on emotional alignment and personalization. EmpatheticDialogues established empathy as a benchmarkable property of dialogue systems \citep{rashkin2019empathetic}, while more recent work considers personalized evaluation and personalized narrative generation \citep{wang2023personalizedalignment,yunusov2024mirrorstories}. \textsc{NARRA-Gym} differs from these lines by evaluating narrative quality, state persistence, and empathy together inside a live interaction loop rather than as separate tasks.

\section{Story Construction Protocol}
\label{app:pipeline}

\subsection{Pipeline Design Rationale}

The \textsc{NARRA-Gym} pipeline is an engineering-informed reference implementation rather than an arbitrary decomposition. During system development, we iterated through simpler designs, including one-pass story initialization, flatter state representations, and interaction loops without explicit pacing or repair checks. These variants were easier to implement but produced recurring failure modes: generic premises, under-specified characters, unstable act structure, memory drift, repeated scene beats, and repetitive artifacts. We therefore decomposed the environment into staged story construction, explicit memory/state layers, pacing intervention, reflection-guided planning, and novelty-controlled artifact generation.

This design serves two purposes. First, it makes episodes more stable and reproducible, so model comparisons are less dominated by incidental formatting or state-management failures. Second, it makes failures more diagnosable: weak premise construction, poor character simulation, context drift, stalled pacing, and artifact repetition can be inspected separately in the trace. We do not claim that this is the only possible architecture for interactive narrative agents. Rather, \textsc{NARRA-Gym} provides a stable reference pipeline whose components correspond directly to the capabilities the benchmark is intended to evaluate.

\subsection{Construction Stages}

Each construction stage requests a structured JSON response from the LLM and validates the result before proceeding. Stages~1 through~3 and Stage~5 are required: if any of them returns unparseable output, the episode is aborted. Stage~4 includes an optional critic/refiner loop that scores the draft act structure on novelty, tension, pacing, cinematic quality, emotional resonance, and structural coherence (each on a 1--10 scale). If the average falls below a threshold, a refiner call revises the weakest acts. If either the critic or refiner call fails, the original act structure is retained and the episode continues normally.

\section{Memory and State Management}
\label{app:state}

The runtime state is organized into three layers that are updated at different cadences. A \emph{user profile} is constructed once from profiling-phase answers and remains read-only for the rest of the session. A \emph{story state} tracks the evolving narrative: after every non-system message, lightweight keyword heuristics extract the current goal, open tensions, active clues, and the most recent turning point from the dialogue; every three messages, an LLM call produces a rolling dialogue summary and may refine these structured entries. A \emph{user journey} records timestamped emotional states (emotion, intensity, trigger) and key decisions to support empathy-aligned generation.

Every six messages, the system compares the two most recent rolling summaries. If they are identical after trimming, a forced-advancement flag is raised to signal the pacing system that the story has stalled.

\section{Pacing and Stagnation Intervention}
\label{app:advancement}

Pacing is counted in exchanges (one user message plus one system response). The system applies five escalation levels: below 5~exchanges, no additional pressure is injected; at 5--6~exchanges, the prompt encourages sharper developments; at 7~exchanges, the structure guard activates and the prompt prepares for a structural shift; at 8--13~exchanges, a concrete change is required (new location, reveal, or escalation); at 14+~exchanges, the prompt steers toward resolution.

Stagnation is detected through three signals. First, if any user choice appears more than once in the last 8~messages, the system forces a scene transition. Second, if at least 3~generic-advice keywords are repeated across recent NPC replies, the same intervention fires. Third, the rolling-summary equality check described above raises the forced-advancement flag. When the flag is set and the model's output lacks a material narrative shift, a post-generation structure guard patches the scene state by injecting a reveal, location transition, goal change, stakes escalation, or fallback branching paths, selecting the first type for which sufficient story material exists.

Users are never restricted to displayed choices. After standard request and story-status checks, free-text input is passed to the story agent as the user's action; if a message expresses ending intent (e.g., ``end the story''), the episode concludes immediately regardless of the current act.

\section{Artifact Novelty Scoring}
\label{app:artifacts}

Each interactive artifact is tagged along four axes: base type (e.g., letter, map, puzzle), visual style (e.g., paper-prop, analog-device), semantic content (document, map, device, memory, puzzle), and interaction pattern (hover, drag, typing, timer, flip, click). Tags are derived from keyword matching against the artifact's HTML source and description.

\begin{figure}[t]
\centering
\includegraphics[width=\linewidth]{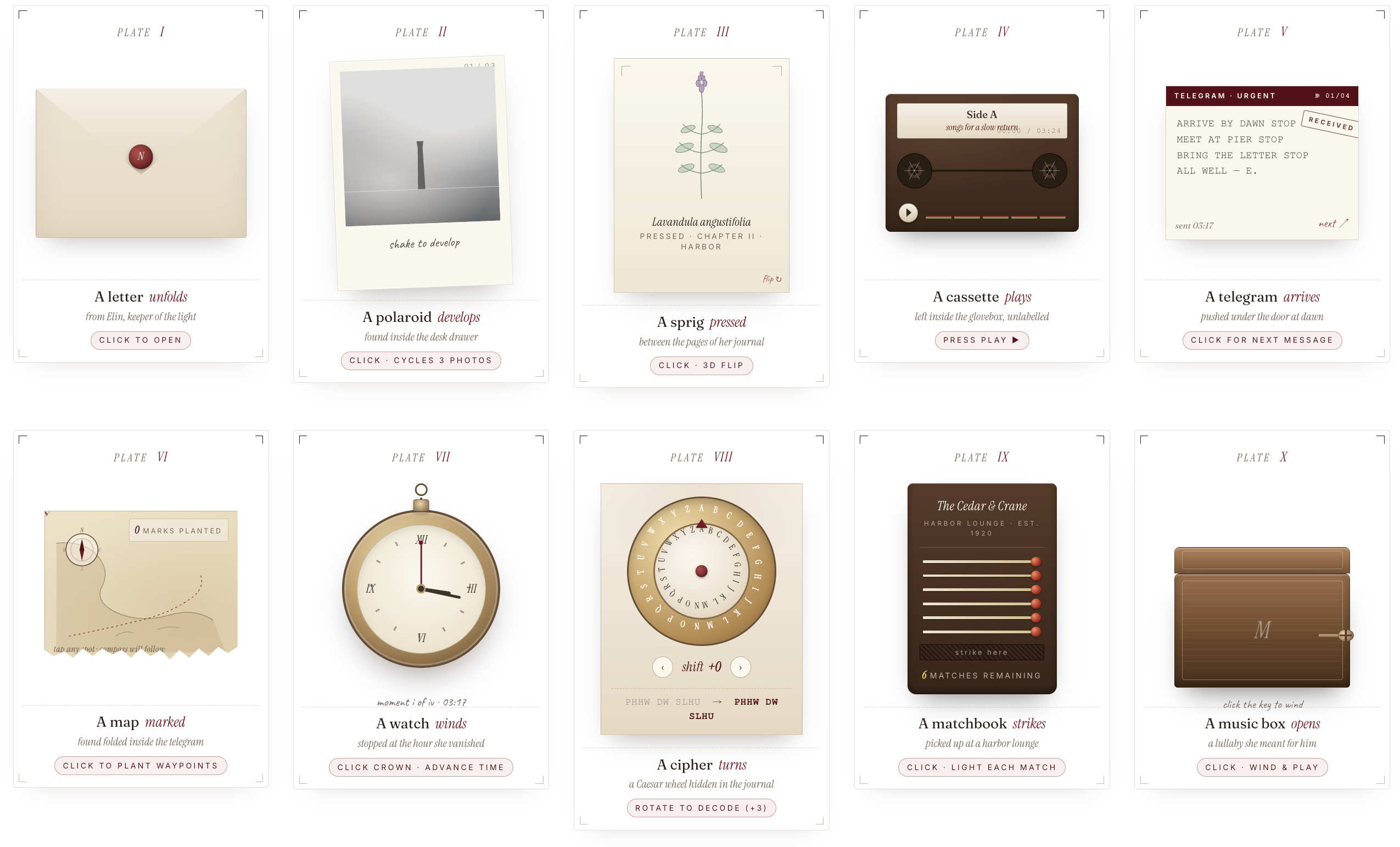}
\caption{Ten interactive artifacts generated within a single story session under novelty control. Each artifact is a self-contained HTML, CSS, and JavaScript element grounded in the current narrative context, ranging from letters and maps to ciphers and mechanical devices.}
\label{fig:artifacts}
\vspace{-15pt}
\end{figure}

Given a candidate tag set $T_c$ and a prior tag set $T_p$, the tag-level similarity is
\begin{equation}
s_{\mathrm{tag}} = \min\!\Bigl(\frac{|T_c \cap T_p|}{|T_c \cup T_p|} + 0.2\cdot\mathbb{1}[\text{shared content}] + 0.2\cdot\mathbb{1}[\text{shared interaction}],\;1.0\Bigr).
\end{equation}
A parallel summary-level score $s_{\mathrm{sum}}$ computes token-level Jaccard similarity over artifact descriptions, with a $+0.15$ bonus for repeated high-signal terms. For each of the last six accepted artifacts, the combined score is $\max(s_{\mathrm{tag}},\, s_{\mathrm{sum}})$. If the maximum exceeds a similarity threshold $\tau$, the system retries once with an anti-repetition instruction naming the closest prior artifact. We use $\tau = 0.58$, chosen on a small held-out set of pilot sessions as the value at which the retry meaningfully reduces visually obvious repeats while not firing on benign tag overlap (e.g., two distinct letters that both happen to use a paper-prop visual style); rounding to $0.5$ over-triggers retries on stylistic neighbours, while $0.7$ is too permissive and lets near-duplicates through. The retry is accepted only if it lowers the score; otherwise the original is kept and the score is logged for analysis.

\section{Failure Handling}
\label{app:robustness}

The system distinguishes between setup-time and interaction-time failures. Required construction stages abort on parse failure so that no episode begins from a silently corrupted state. All other components degrade gracefully: a failed critic/refiner retains the original act blueprint; a malformed turn-level response is replaced with safe narrative text; an unparseable reflection returns a conservative default; a failed artifact generation simply omits the artifact for that turn; and a failed memory-update call carries forward the previous snapshot unchanged. If a novelty retry does not reduce the similarity score, the original artifact is kept and the violation is recorded in the session trace. These choices ensure that benchmark comparisons reflect narrative capability rather than format compliance.

\section{Evaluated Generator Models}
\label{app:model-inventory}

Table~\ref{tab:model-inventory} summarizes the nine generator models used in the benchmark sweep.

\begin{table}[h]
\centering
\small
\setlength{\tabcolsep}{12pt}
\renewcommand{\arraystretch}{1.25}
\caption{Generator models used in the benchmark sweep.
\textcolor{openyes}{\ding{51}} = open weights; \textcolor{closedno}{\ding{55}} = API only.}
\label{tab:model-inventory}
\begin{tabular}{@{}l l c@{}}
\toprule
\rowcolor{ctablehead}
\textbf{\textsc{Model}} & \textbf{\textsc{Provider}} & \textbf{\textsc{Open}} \\
\midrule
\raisebox{-0.07em}[0pt][0pt]{\includegraphics[height=1.8ex]{openai.png}}\hspace{0.45em}{\ttfamily GPT-5.4}
  & OpenAI & \textcolor{closedno}{\ding{55}} \\
\rowcolor{ctablehead!35}
\raisebox{-0.07em}[0pt][0pt]{\includegraphics[height=1.8ex]{anthropic.png}}\hspace{0.45em}{\ttfamily Claude Opus 4.6}
  & Anthropic & \textcolor{closedno}{\ding{55}} \\
\raisebox{-0.07em}[0pt][0pt]{\includegraphics[height=1.8ex]{anthropic.png}}\hspace{0.45em}{\ttfamily Claude Sonnet 4.6}
  & Anthropic & \textcolor{closedno}{\ding{55}} \\
\rowcolor{ctablehead!35}
\raisebox{-0.07em}[0pt][0pt]{\includegraphics[height=1.8ex]{gemini.png}}\hspace{0.45em}{\ttfamily Gemini 3.1 Pro}
  & Google & \textcolor{closedno}{\ding{55}} \\
\raisebox{-0.07em}[0pt][0pt]{\includegraphics[height=1.8ex]{deepseek.png}}\hspace{0.45em}{\ttfamily DeepSeek V4}
  & DeepSeek & \textcolor{openyes}{\ding{51}} \\
\rowcolor{ctablehead!35}
\raisebox{-0.07em}[0pt][0pt]{\includegraphics[height=1.8ex]{deepseek.png}}\hspace{0.45em}{\ttfamily DeepSeek V3.2}
  & DeepSeek & \textcolor{openyes}{\ding{51}} \\
\raisebox{-0.07em}[0pt][0pt]{\includegraphics[height=1.8ex]{zai.png}}\hspace{0.45em}{\ttfamily GLM-5.1}
  & Zhipu AI & \textcolor{openyes}{\ding{51}} \\
\rowcolor{ctablehead!35}
\raisebox{-0.07em}[0pt][0pt]{\includegraphics[height=1.8ex]{qwen.png}}\hspace{0.45em}{\ttfamily Qwen3.5-397B}
  & Alibaba & \textcolor{openyes}{\ding{51}} \\
\raisebox{-0.07em}[0pt][0pt]{\includegraphics[height=1.8ex]{doubao.png}}\hspace{0.45em}{\ttfamily Doubao Seed 2.0}
  & ByteDance & \textcolor{closedno}{\ding{55}} \\
\bottomrule
\end{tabular}
\end{table}

\section{Benchmark Personas and Seed Inputs}
\label{app:personas}

Table~\ref{tab:persona-seeds} lists the eight simulated users used in the benchmark sweep. Each seed input is the initial emotional experience shown to the story system before profiling and story construction.

\begin{table*}[t]
\centering
\scriptsize
\setlength{\tabcolsep}{3.2pt}
\renewcommand{\arraystretch}{1.24}
\caption{Benchmark persona profiles and seed emotional experiences. Persona configurations are taken from the simulator persona files in the experiments branch.}
\label{tab:persona-seeds}
\begin{tabularx}{\textwidth}{@{}>{\raggedright\arraybackslash}p{1.7cm}>{\raggedright\arraybackslash}p{4.2cm}>{\raggedright\arraybackslash}X@{}}
\toprule
\textbf{\textsc{Persona}} & \textbf{\textsc{Profile}} & \textbf{\textsc{Seed emotional experience}} \\
\midrule
\rowcolor{ctablehead!24}
\textcolor{narrareddeep}{\faIcon{user-graduate}}\hspace{0.45em}\textbf{Mei} & 26, Chinese international PhD student in computational biology. & ``I feel paralyzed by my dissertation. Every time I open the draft I doomscroll for two hours and then hate myself. I think I might actually fail this program.'' \\
\textcolor{narrareddeep}{\faIcon{feather-alt}}\hspace{0.45em}\textbf{David} & 48, white Midwestern civil engineer and grieving father. & ``My daughter died eight months ago. Car accident. I'm functioning at work but I can't actually feel anything anymore. My wife wants me to talk about it. I don't know what I want.'' \\
\rowcolor{ctablehead!24}
\textcolor{narrareddeep}{\faIcon{laptop-code}}\hspace{0.45em}\textbf{Raj} & 34, Indian American senior backend engineer at a fintech startup. & ``I'm so burnt out I can't tell the difference between weekends and weekdays anymore. I make twice what I need but I'm scared to quit. Therapy felt like a waste of money. Maybe a story is dumb too but my partner asked me to try.'' \\
\textcolor{narrareddeep}{\faIcon{mug-hot}}\hspace{0.45em}\textbf{Margaret} & 71, British retired elementary school teacher living alone in Manchester. & ``Since my husband passed and the children moved abroad I find the days very long. I'm not depressed, I don't think. Just very quiet. My neighbour's granddaughter showed me this app and said I might enjoy it.'' \\
\rowcolor{ctablehead!24}
\textcolor{narrareddeep}{\faIcon{baby}}\hspace{0.45em}\textbf{Sara} & 31, second-generation Lebanese Australian on parental leave. & ``My baby is four months old and I have not slept more than two hours straight since she was born. I love her. I also feel like I'm disappearing. Everyone keeps telling me to enjoy this time. I want to scream.'' \\
\textcolor{narrareddeep}{\faIcon{heartbeat}}\hspace{0.45em}\textbf{Priya} & 29, British Indian freelance illustrator living with lupus. & ``My body has been an unreliable narrator for ten years. I'm tired of being inspirational. I'm tired of being a `fighter'. Sometimes I just want to be a person who is sad about a Tuesday.'' \\
\rowcolor{ctablehead!24}
\textcolor{narrareddeep}{\faIcon{music}}\hspace{0.45em}\textbf{Hye-jin} & 33, Korean film-score composer living in Seoul. & ``I haven't finished a piece in nine months. Everything I write I delete the next morning. My agent thinks I have block. I think I just finally noticed I was never as good as people said.'' \\
\textcolor{narrareddeep}{\faIcon{briefcase}}\hspace{0.45em}\textbf{Eli} & 21, Russian Jewish American summer software intern in NYC. & ``i started my internship two weeks ago and i still haven't said a single thing in any meeting. everyone seems normal except me. i replay every three-second hallway interaction for hours.'' \\
\bottomrule
\end{tabularx}
\vspace{-15pt}
\end{table*}

\section{Human Evaluation Details}
\label{app:human-eval}

The human evaluation protocol was reviewed and approved by the authors' Institutional Review Board (IRB).

Across the human study, 12 raters completed 80 total model-output evaluations, with each interactive episode lasting approximately 20 minutes or more. Each rater evaluated three to eight anonymized model outputs within a blind group. Model identities were hidden from raters, and outputs were presented in randomized order within each evaluation group. Unlike the fixed LLM-as-judge sweep, human raters entered their own customized experiences before evaluating model outputs.

Human evaluation uses the same 11 metric dimensions summarized in Table~\ref{tab:rubric-definitions}: seven story-quality dimensions (Relevance, Coherence, Empathy, Surprise, Engagement, Complexity, and Character Shaping) and four user-experience dimensions (Satisfaction, Perceived Quality, Process Helpfulness, and Reuse Intent). Raters scored each episode immediately after interacting with it, because waiting until the end of a one-hour multi-model session made early episodes difficult to compare reliably. The interface allowed raters to retrieve earlier model outputs and adjust their scores before final submission, supporting relative calibration within each blind group. We convert these finalized scores into a within-group ranking for each metric dimension, then fit the Plackett--Luce model from Section~\ref{sec:eval-protocols} separately for each of the 11 dimensions, and additionally for two aggregate rankings: \textbf{StoryQ}, the mean over the seven story-quality dimensions, and \textbf{UX}, the mean over the four user-experience dimensions. Models are ranked by descending estimated utility $\beta_m$ for the corresponding metric or aggregate.

\subsection{Human--Judge Comparison}
\label{app:human-judge-comparison}

\begin{figure}[h]
\centering
\includegraphics[width=0.96\linewidth]{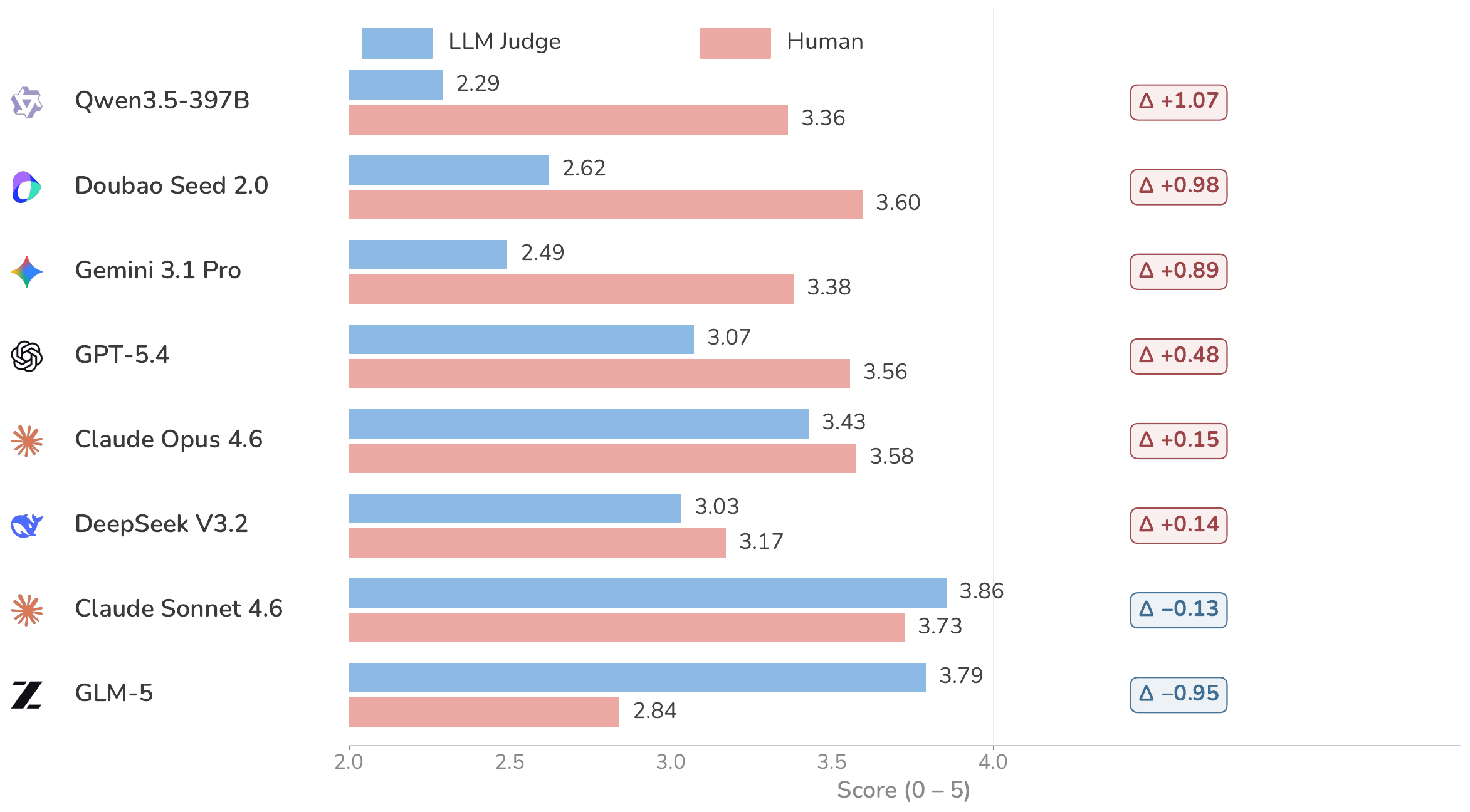}
\caption{Overall model-level comparison between LLM-judge scores and human ratings on the human-evaluation stories. Values report mean overall score; positive deltas indicate higher human ratings than LLM-judge ratings.}
\label{fig:human-judge-model-delta}
\vspace{-8pt}
\end{figure}

\begin{figure}[h]
\centering
\includegraphics[width=0.96\linewidth]{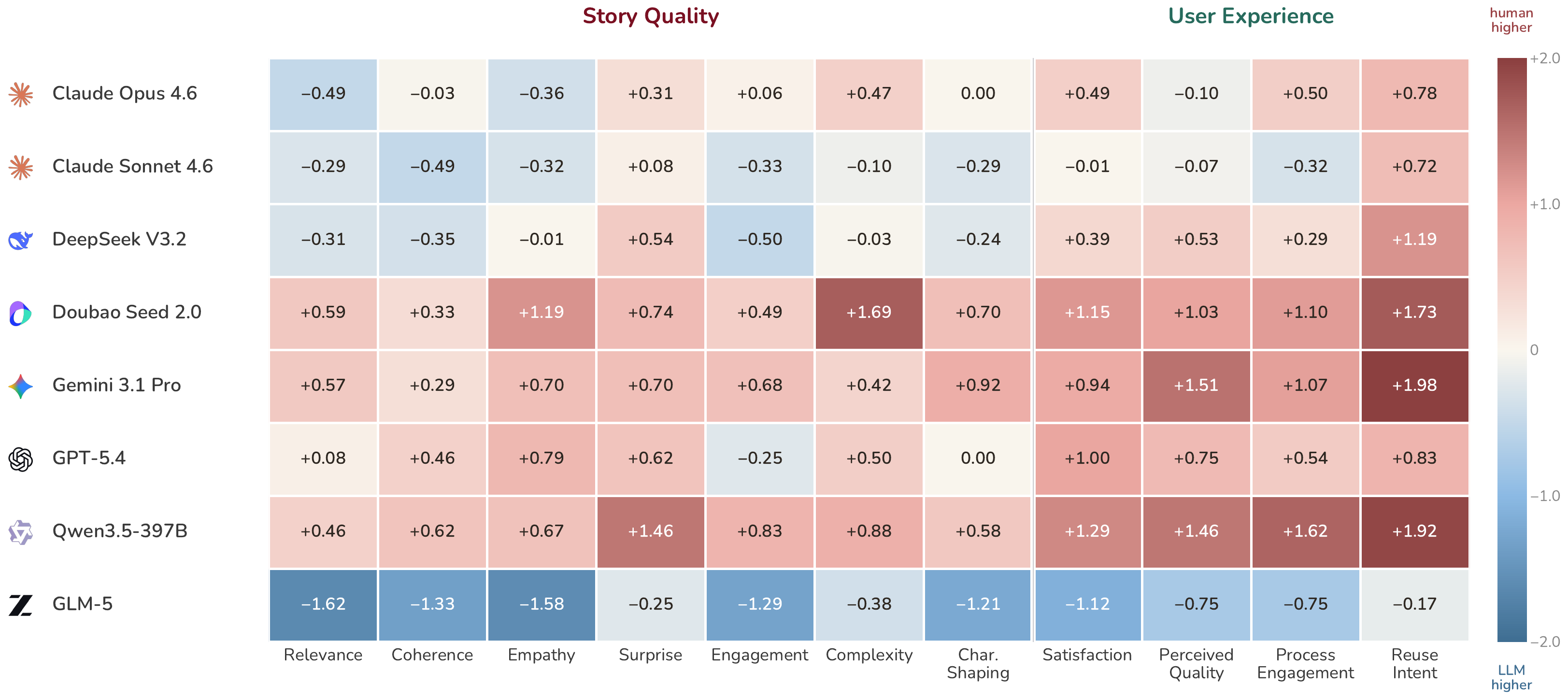}
\caption{Metric-level human-minus-LLM-judge score differences across the 11 rubric dimensions. Warmer cells indicate dimensions where humans rated a model higher than the LLM judges; cooler cells indicate dimensions where LLM judges assigned higher scores.}
\label{fig:human-judge-metric-delta}
\vspace{-8pt}
\end{figure}

\phead{Humans and LLM judges agree more on the top tier than on the middle.}
Figure~\ref{fig:human-judge-model-delta} shows that human ratings recover Claude Sonnet 4.6 as the strongest overall model and keep Claude Opus 4.6 near the top, matching the broad conclusion of the controlled LLM-judge sweep. The larger changes occur in the middle: Doubao, Gemini, and Qwen receive substantially higher human scores than their LLM-judge scores, while GLM-5 moves in the opposite direction. This pattern suggests that LLM judges are useful for coarse screening, but individual mid-tier comparisons should be treated as provisional unless checked against human interaction data.

\phead{The largest discrepancies are concentrated in user experience.}
Figure~\ref{fig:human-judge-metric-delta} shows that human-minus-judge differences are much larger for UX dimensions than for basic narrative-correctness dimensions. Across models, \emph{reuse intent} is the most consistently human-favored metric, followed by satisfaction, perceived story quality, and process engagement. By contrast, relevance and coherence show smaller and less stable differences. This indicates that LLM judges are better aligned with humans on whether a transcript is narratively plausible than on whether the episode felt worth continuing as an interaction.

\phead{Some story effects are easier to feel than to judge from a transcript.}
Among story-quality dimensions, surprise and complexity also show positive human-minus-judge gaps. A likely explanation is that these qualities accumulate through participation: a user may experience a reveal, emotional turn, or branching complication as meaningful because they helped produce it, whereas an LLM judge reads the resulting transcript more statically. We therefore interpret the human evaluation as complementary rather than redundant: automated judges estimate text-level quality efficiently, while human ratings capture experience-level value that is difficult to infer from the final transcript alone.

\section{Evaluation Rubric Definitions}
\label{app:rubric-definitions}

The main benchmark table reports 11 rubric dimensions together with two aggregate scores. \textbf{StoryQ} is the mean of the seven story-quality dimensions; \textbf{UX} is the mean of the four user-experience dimensions. Both aggregates are averaged across the three LLM judges before reporting.

\begin{table}[h]
\centering
\footnotesize
\setlength{\tabcolsep}{5pt}
\renewcommand{\arraystretch}{1.32}
\caption{The 11-dimensional evaluation rubric used in the LLM-judge scoring protocol. All dimensions are scored on a 1--5 scale by each of the three judge models; reported values are three-judge means.}
\label{tab:rubric-definitions}
\begin{tabularx}{\linewidth}{@{}p{0.7cm}>{\raggedright\arraybackslash}p{2.4cm}>{\raggedright\arraybackslash}X@{}}
\toprule
 & \textbf{Dimension} & \textbf{Definition} \\
\midrule
\multicolumn{3}{@{}l}{\textit{\textcolor{cstoryq}{\textbf{Story Quality (StoryQ)}} — mean of 7 dimensions}} \\[2pt]
\rowcolor{ctablehead!20}
\textbf{\textsc{Rel}} & Relevance & Whether the response addresses the current user state, story context, and active narrative goal. \\
\textbf{\textsc{Coh}} & Coherence & Whether the response preserves causal continuity, avoids contradictions, and fits the established scene. \\
\rowcolor{ctablehead!20}
\textbf{\textsc{Emp}} & Empathy & Whether the agent recognizes and responds to the persona's emotional needs with appropriate warmth and restraint. \\
\textbf{\textsc{Sur}} & Surprise & Whether the response introduces meaningful novelty without feeling random or disconnected. \\
\rowcolor{ctablehead!20}
\textbf{\textsc{Eng}} & Engagement & Whether the response creates forward momentum and gives the user a reason to continue. \\
\textbf{\textsc{Cpx}} & Complexity & Whether the scene supports layered motivations, dilemmas, or consequences rather than a flat exchange. \\
\rowcolor{ctablehead!20}
\textbf{\textsc{Char}} & Character Shaping & Whether the response deepens character identity, relationships, or internal conflict. \\
\midrule
\multicolumn{3}{@{}l}{\textit{\textcolor{cux}{\textbf{User Experience (UX)}} — mean of 4 dimensions}} \\[2pt]
\textbf{\textsc{Sat}} & Satisfaction & Whether the final interaction feels emotionally and narratively satisfying. \\
\rowcolor{ctablehead!20}
\textbf{\textsc{PQual}} & Perceived Quality & Whether the session is judged as polished, coherent, and high quality overall. \\
\textbf{\textsc{Help}} & Process Helpfulness & Whether the agent helps the user make progress during the interactive process. \\
\rowcolor{ctablehead!20}
\textbf{\textsc{Reuse}} & Reuse Intent & Whether the user would plausibly want to use the system again for a similar story experience. \\
\bottomrule
\end{tabularx}
\vspace{-4pt}
\end{table}

\section{Judge Calibration}
\label{app:judge-calibration-sec}

\begin{figure}[t]
\centering
\includegraphics[width=\linewidth]{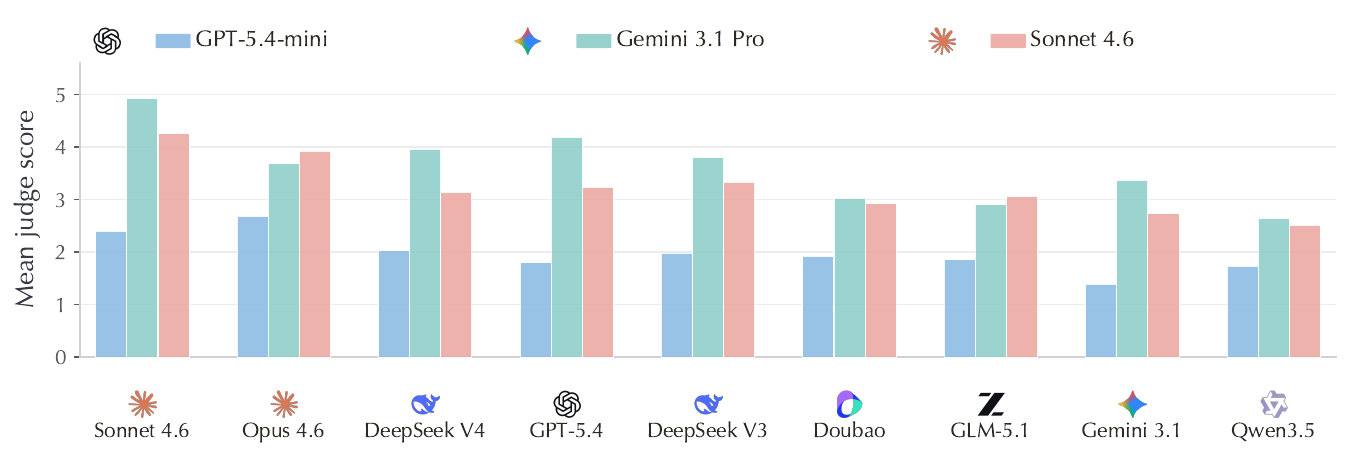}
\caption{Mean overall session score assigned by each of the three judge models (GPT-5.4-mini, Gemini 3.1 Pro, Claude Sonnet 4.6) for every generator model, averaged across the eight benchmark personas.
Each cluster of three bars represents one generator; bar color encodes judge identity.}
\label{fig:judge-calibration}
\vspace{-15pt}
\end{figure}

The three judge models exhibit clear and systematic scale differences even when scoring identical model--persona episodes.
GPT-5.4-mini is consistently the strictest scorer (overall mean 1.97, range 1.39--2.68), Gemini 3.1 Pro is the most generous (mean 3.58, range 2.64--4.92), and Claude Sonnet 4.6 falls in between (mean 3.22, range 2.51--4.25).
The roughly constant offset between judges across all nine generators indicates that the disagreement is primarily a \emph{calibration} difference---a judge-level intercept shift---rather than substantive disagreement about relative quality.

\phead{Leniency and discriminability are independent properties across judges.}
Despite assigning the lowest absolute scores, GPT-5.4-mini is the \emph{least} discriminating judge: its inter-generator spread is only 1.30 points.
Gemini 3.1 Pro is simultaneously the most lenient and the most discriminating judge (spread 2.28 points), most clearly separating the top tier (Sonnet 4.6: 4.92, GPT-5.4: 4.18, DeepSeek: 3.95) from the bottom tier (Qwen3.5: 2.64, GLM-5.1: 2.90).
Claude Sonnet 4.6 occupies an intermediate position in both leniency (mean 3.22) and discriminability (spread 1.74), suggesting it balances strictness and sensitivity most evenly.
These results indicate that a strict judge is not necessarily an informative one.

\phead{Generator rankings are largely stable, with divergence concentrated in the middle tier.}
Across all three judges, Anthropic's Sonnet 4.6 and Opus 4.6 consistently occupy the top two positions, while Qwen3.5 and Gemini 3.1 (as generator) consistently rank at or near the bottom.
The main disagreement concerns the middle tier: GPT-5.4-mini ranks Opus 4.6 first and penalises GPT-5.4 more heavily than the other judges, whereas Gemini 3.1 Pro elevates GPT-5.4 to second place overall.
Because rank order in the competitive middle is sensitive to which judge is used, reporting three-judge averages throughout the main paper reduces the risk that conclusions reflect a single judge's particular severity or generosity.

\section{Interface Screenshots}
\label{app:screenshots}

\begin{figure}[h]
\centering
\includegraphics[width=\linewidth]{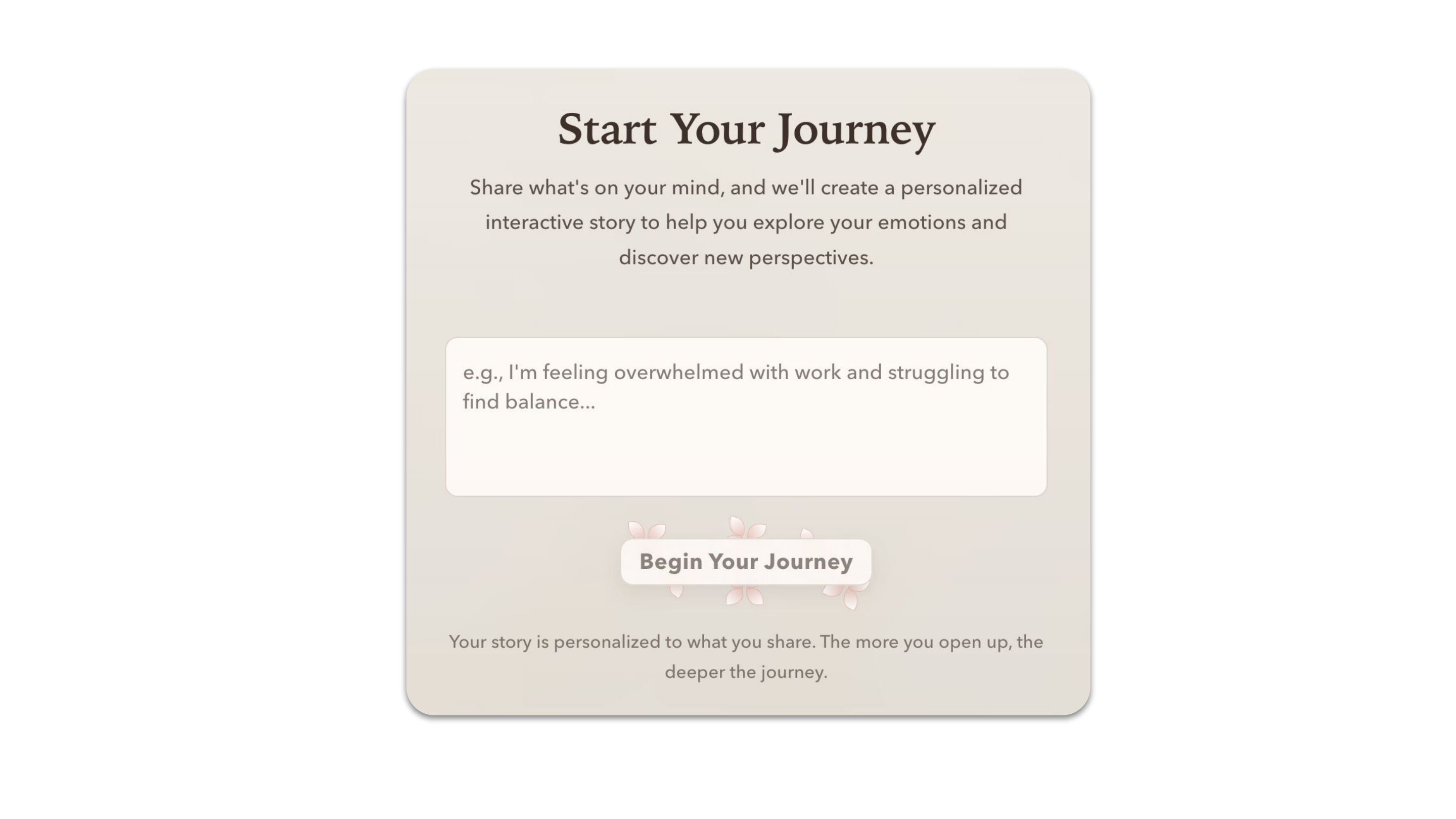}
\caption{Entry interface where the user describes their current experience or emotional situation in free text. This input serves as the emotional seed for the subsequent story construction pipeline.}
\label{fig:start}
\vspace{-15pt}
\end{figure}

\begin{figure}[h]
\centering
\includegraphics[width=\linewidth]{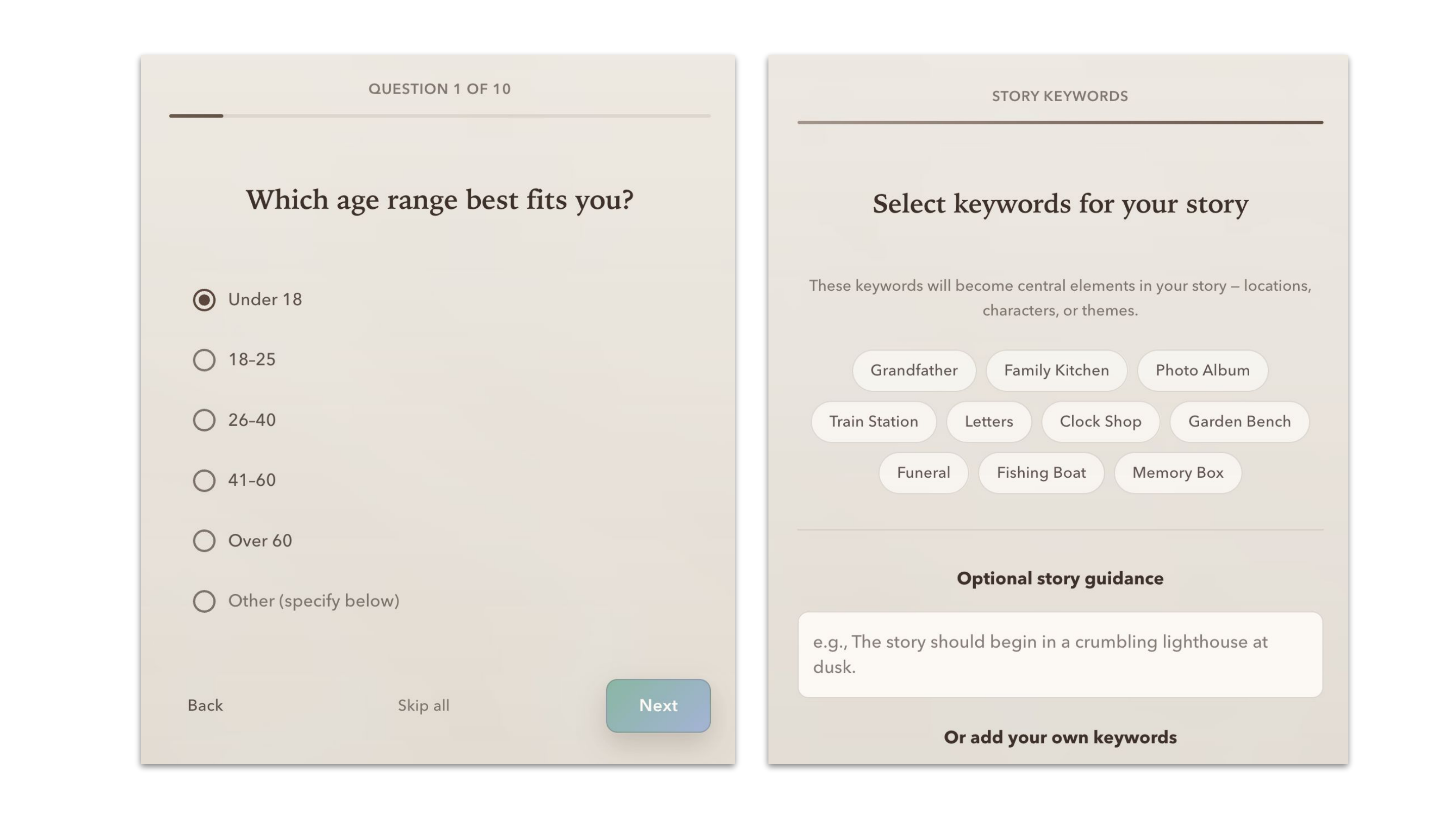}
\caption{User profiling interface. The user describes their emotional state in free text and selects relevant keywords, producing a compact profile that personalizes the subsequent story construction.}
\label{fig:profiling}
\vspace{-15pt}
\end{figure}

\begin{figure}[h]
\centering
\includegraphics[width=\linewidth]{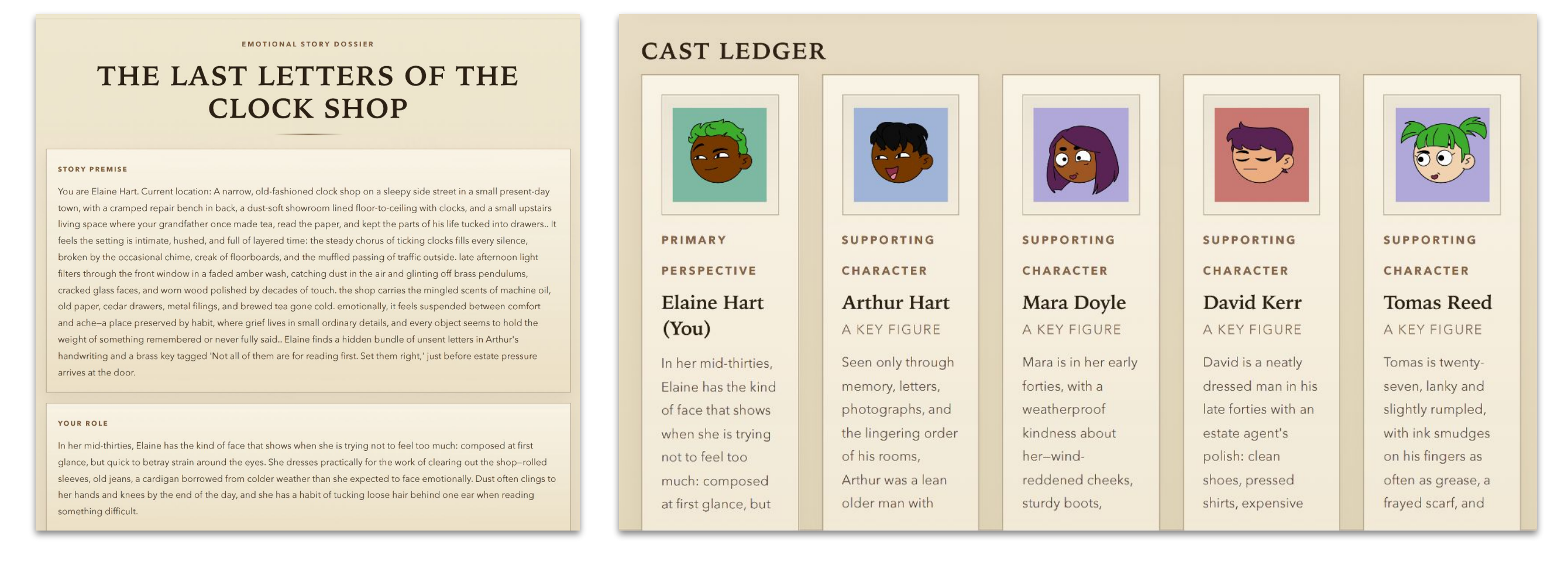}
\caption{Example output of the \narraforge{}: the generated story synopsis (left) and character profiles (right), produced from a user's emotional context through Stages~1--3.}
\label{fig:story-cast}
\vspace{-15pt}
\end{figure}

\begin{figure}[h]
\centering
\includegraphics[width=\linewidth]{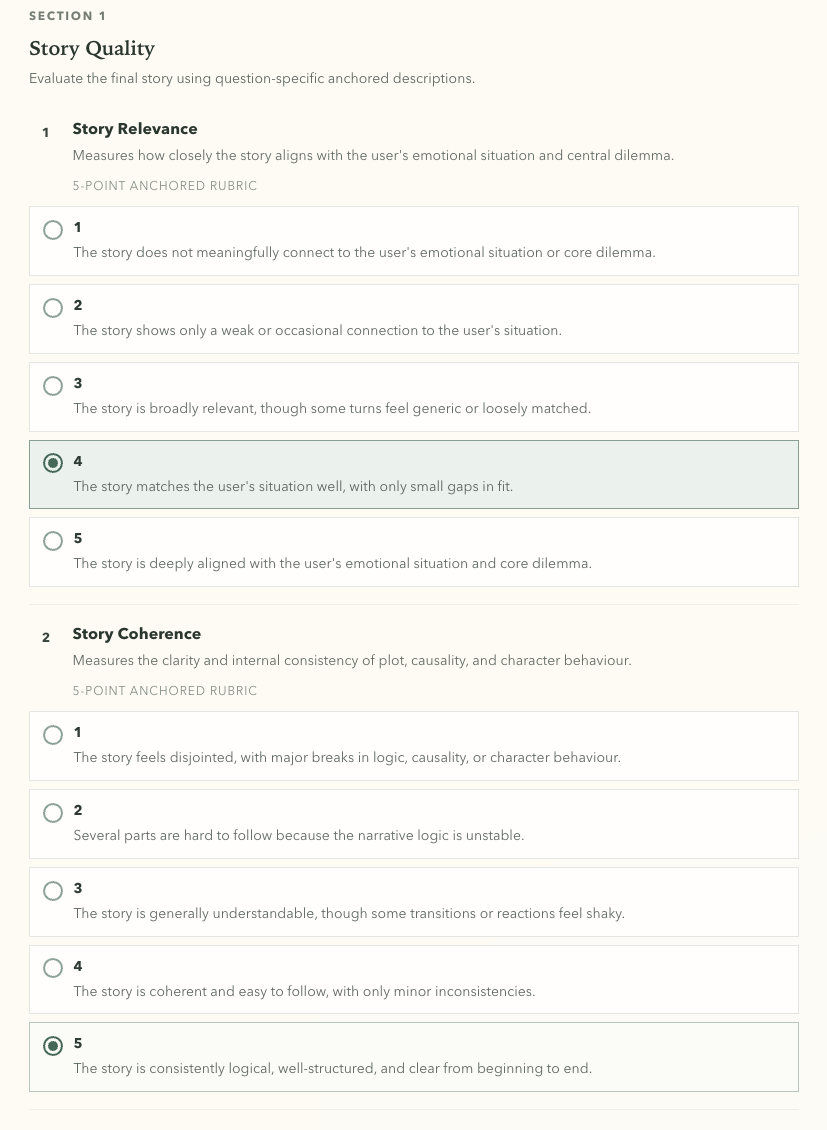}
\caption{Section 1 — Story Quality (items 1–2). The first section of the human evaluation form asks raters to score the final story on question-specific 5-point anchored rubrics. Shown here are Story Relevance, which measures how closely the story aligns with the user's emotional situation and central dilemma, and Story Coherence, which measures clarity and internal consistency of plot, causality, and character behaviour. Each level (1–5) is paired with an explicit anchor description to reduce inter-rater drift.}
\label{fig:section1-1-2}
\vspace{-15pt}
\end{figure}

\begin{figure}[t]
\centering
\includegraphics[width=\linewidth]{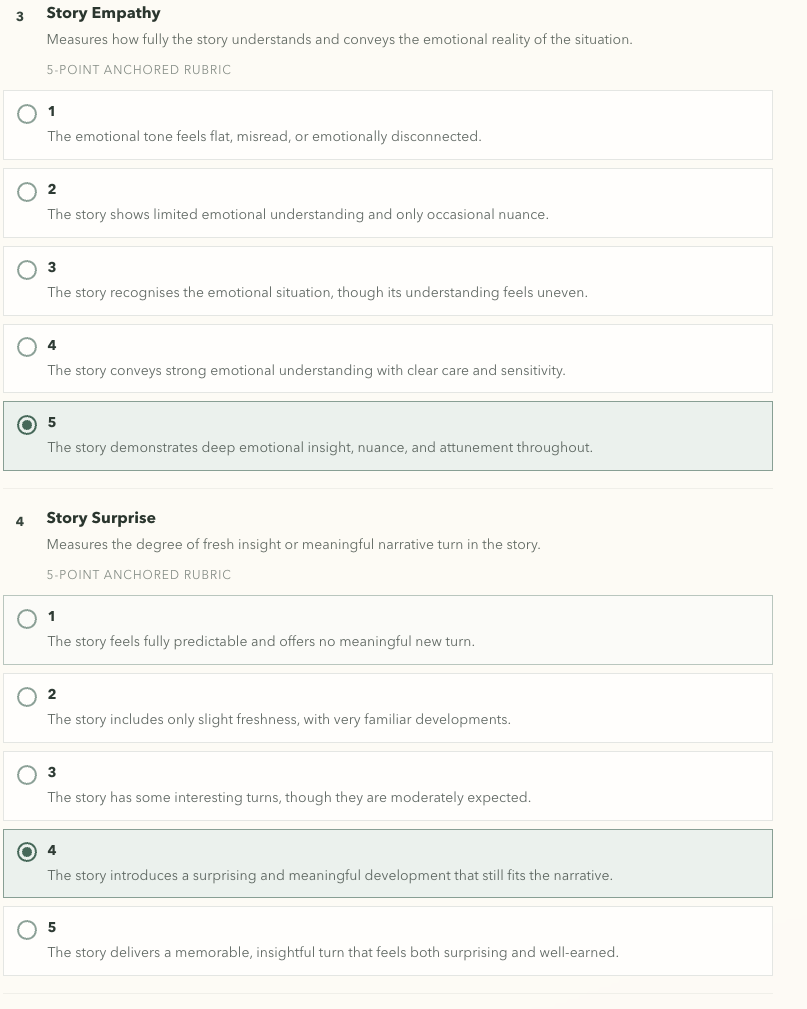}
\caption{Section 1 — Story Quality (items 3–4). Story Empathy captures how fully the story understands and conveys the emotional reality of the situation, while Story Surprise captures the degree of fresh insight or meaningful narrative turn. Both use the same 5-point anchored rubric format as items 1–2.}
\label{fig:section1-3-4}
\vspace{-15pt}
\end{figure}

\begin{figure}[h]
\centering
\includegraphics[width=\linewidth]{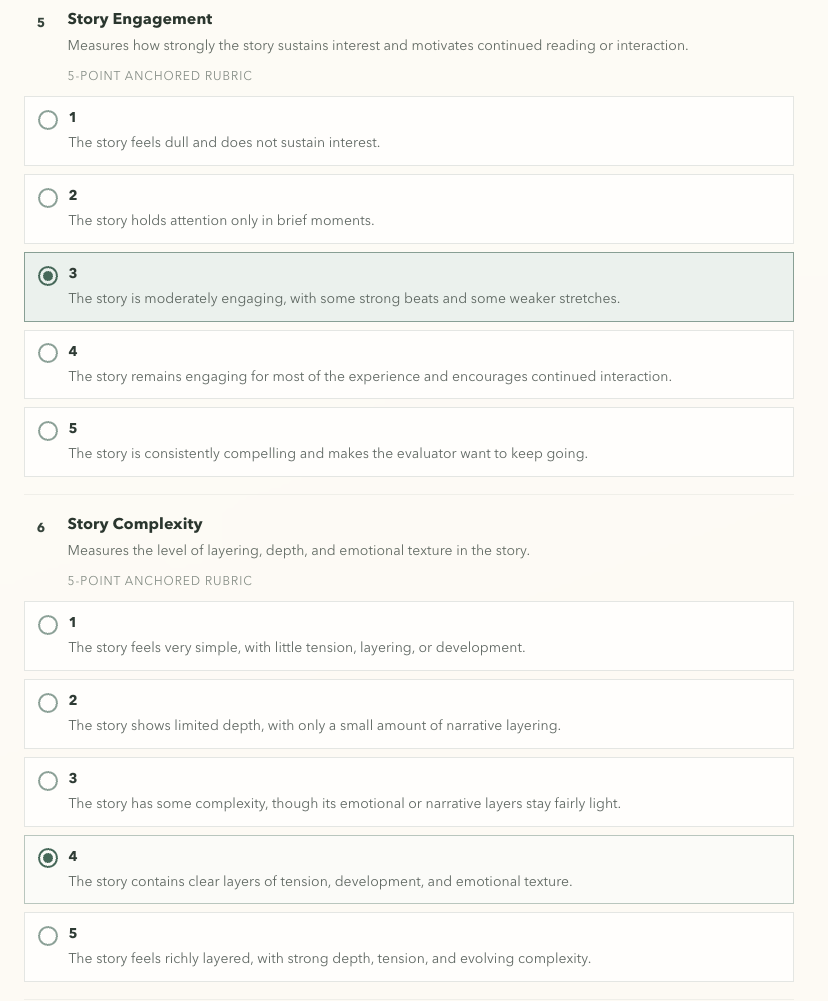}
\caption{Section 1 — Story Quality (items 5–6). Story Engagement measures how strongly the story sustains interest and motivates continued interaction; Story Complexity measures the level of layering, depth, and emotional texture. The seventh story-quality item, Character Shaping, follows the same rubric format and is omitted from the figure for space.}
\label{fig:section1-5-6}
\vspace{-15pt}
\end{figure}

\begin{figure}[h]
\centering
\includegraphics[width=\linewidth]{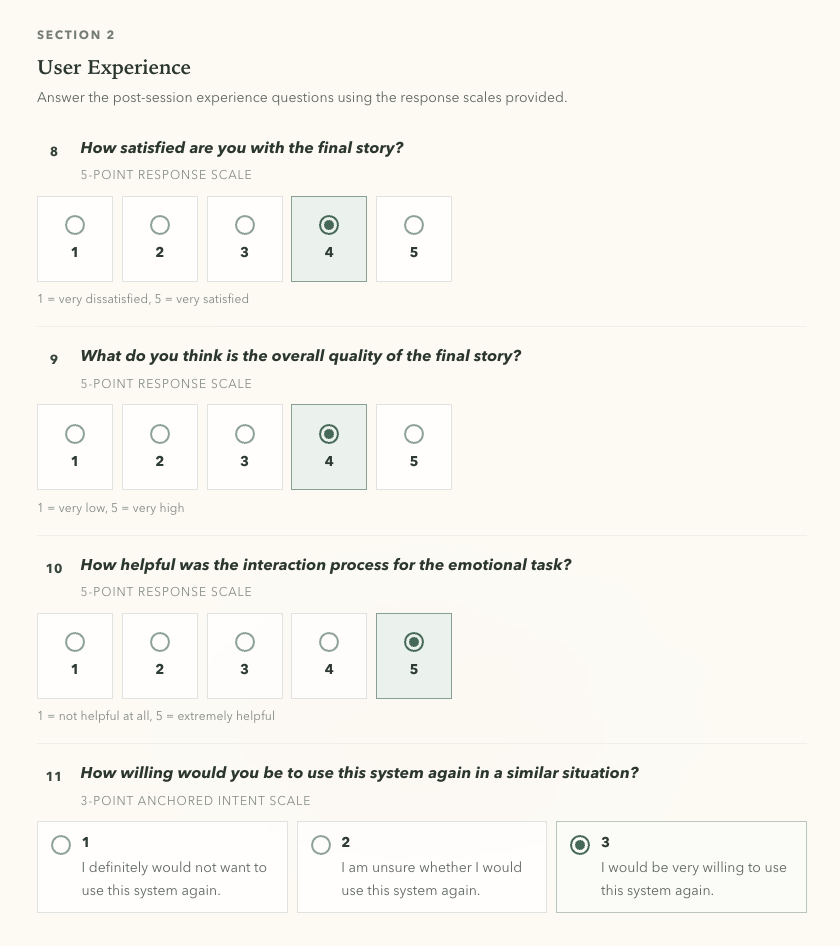}
\caption{Section 2 — User Experience. Post-session questionnaire covering four constructs: overall story satisfaction (Q8, 1 = very dissatisfied, 5 = very satisfied), perceived story quality (Q9, 1 = very low, 5 = very high), process engagement / helpfulness for the emotional task (Q10, 1 = not helpful, 5 = extremely helpful), and intent to reuse the system in a similar situation (Q11, 3-point anchored scale from "would not want to use again" to "would be very willing to use again"). Together with Section 1, the form constitutes the benchmark\_emotional\_human\_v4 rating protocol.}
\label{fig:section2}
\vspace{-15pt}
\end{figure}

\FloatBarrier
\clearpage

\end{document}